\documentclass[10pt,twocolumn,letterpaper]{article}
\usepackage{iccv}
\usepackage{times}
\usepackage{epsfig}
\usepackage{graphicx}
\usepackage{amsmath}
\usepackage{amssymb}
\usepackage{microtype}

\usepackage{booktabs}
\usepackage[caption=false,font=footnotesize]{subfig}
\usepackage{pifont}
\usepackage{relsize}
\usepackage{multirow}
\usepackage{gensymb}
\usepackage{tabularx}
\usepackage{comment}

\usepackage{etoolbox}
\makeatletter
\patchcmd\@combinedblfloats{\box\@outputbox}{\unvbox\@outputbox}{}{
   \errmessage{\noexpand\@combinedblfloats could not be patched}
}
\makeatother

\usepackage{enumitem}
\setitemize[0]{leftmargin=10pt}

\newenvironment{tight_itemize}{
\begin{itemize}[leftmargin=10pt]
  \setlength{\topsep}{0pt}
  \setlength{\itemsep}{2pt}
  \setlength{\parskip}{0pt}
  \setlength{\parsep}{0pt}
}{\end{itemize}}

\usepackage[pagebackref=true,breaklinks=true,letterpaper=true,colorlinks,bookmarks=false]{hyperref}

\iccvfinalcopy

\def\iccvPaperID{2315}

\ificcvfinal\fi
\begin{document}

\title{Pose-variant 3D Facial Attribute Generation}

\author{Feng-Ju Chang$^{\dagger}$, Xiang Yu$^{\ddag}$, Ram Nevatia$^{\dagger}$, and Manmohan Chandraker$^{\S\ddag}$\\
$^{\dagger}$University of Southern California \\
$^{\S}$University of California, San Diego \\
$^{\ddag}$ NEC Laboratories America\\
{\tt\small \{fengjuch,nevatia\}@usc.edu, \{xiangyu,manu\}@nec-labs.com}}

\maketitle

\begin{abstract}
We address the challenging problem of generating facial attributes using a single image in an unconstrained pose. In contrast to prior works that largely consider generation on 2D near-frontal images, we propose a GAN-based framework to generate attributes directly on a dense 3D representation given by UV texture and position maps, resulting in photorealistic, geometrically-consistent and identity-preserving outputs. Starting from a self-occluded UV texture map obtained by applying an off-the-shelf 3D reconstruction method, we propose two novel components. First, a texture completion generative adversarial network (TC-GAN) completes the partial UV texture map. Second, a 3D attribute generation GAN (3DA-GAN) synthesizes the target attribute while obtaining an appearance consistent with 3D face geometry and preserving identity. Extensive experiments on CelebA, LFW and IJB-A show that our method achieves consistently better attribute generation accuracy than prior methods, a higher degree of qualitative photorealism and preserves face identity information.
\end{abstract}

\section{Introduction}
Faces are of unique interest in computer vision, whether it be for recognition, visualization or animation, upon the diversity with which their images are manifested. This is partly due to the variety of attributes associated with faces and partly due to extrinsic variations like head pose. Thus, generating photorealistic images of faces that address both of those aspects is a problem of fundamental interest that also enables downstream applications, such as augmentation of under-represented classes in face recognition.

\begin{figure}[t]
\centering
\includegraphics[width=0.47\textwidth]{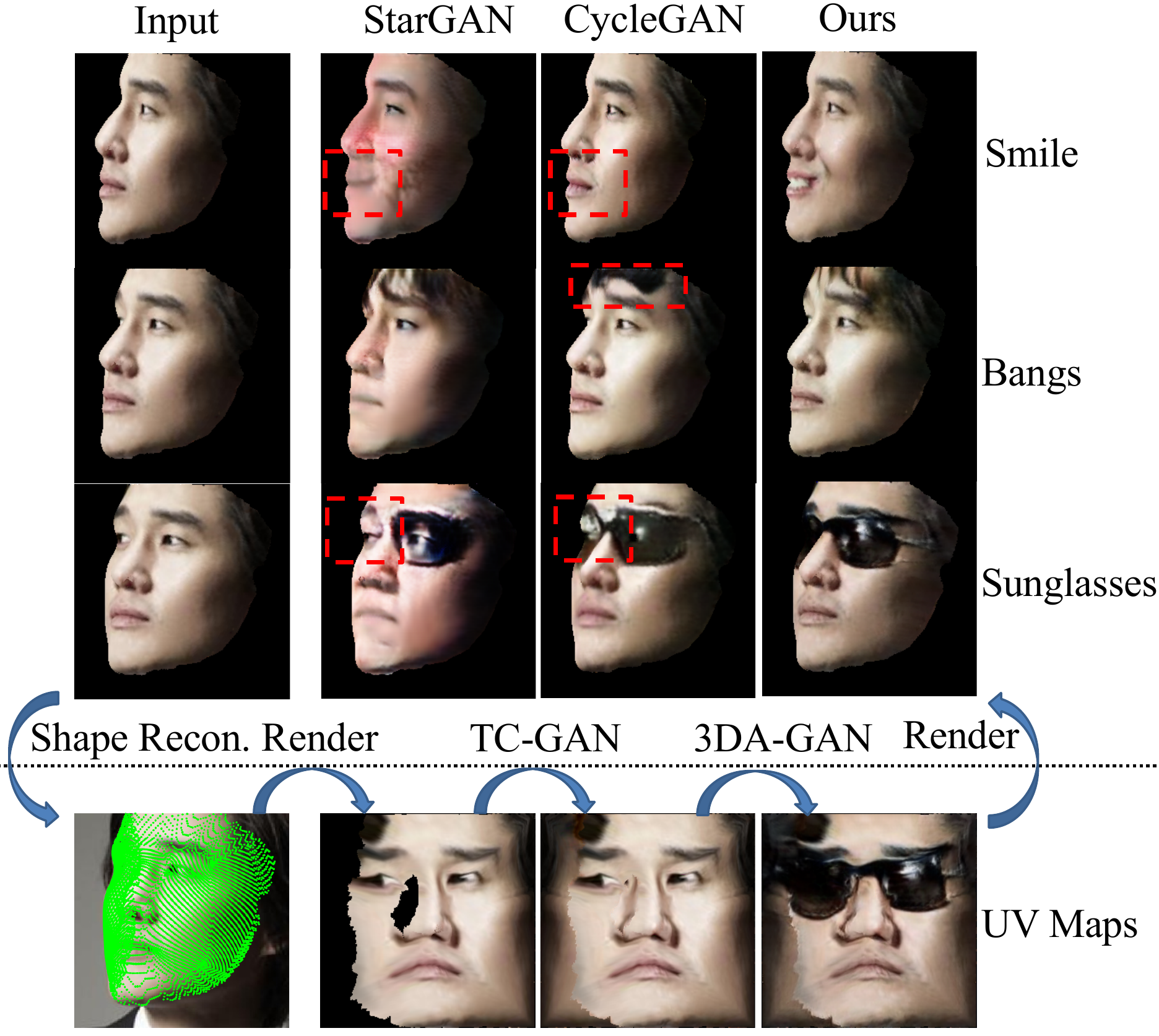}
\caption{Facial attributes generation under head pose variations, showing results comparison of our method to StarGAN~\cite{stargan} and CycleGAN~\cite{CycleGAN2017}. Traditional frameworks generate artifacts due to pose variations. Introducing a 3D UV representation, the proposed TC-GAN and 3DA-GAN generates photo-realistic face attributes on pose-variant faces.}
\vspace{-5mm}
\label{fig:teasor}
\end{figure}

In recent years, conditional generative models such as Variational Auto-Encodesr (VAE)~\cite{vae2014} or Generative Adversarial Networks (GAN)~\cite{gan2014} have achieved impressive results  \cite{yan_eccv2016,stargan,attgan,shen2017}. However, they have largely focused on frontal faces. In contrast, we consider the problem of generating 3D-consistent attributes on possibly pose-variant faces. As a motivating example, consider the problem of adding sunglasses to a face image. For a frontal input and with a desired frontal output, this involves inpainting with sunglass texture limited to the region around the eyes. For an input face image observed under a largely profile view and a more general task of generating an identity-preserving and sunglass-augmented face under arbitrary pose, a more complex transformation is needed since (i) both attribute-related and unrelated regions must be handled and (ii) the attribute must be consistent with 3D face geometry. Technically, this requires working with a higher-dimensional output space and generating an image conditioned on both head pose and attribute code. In Figure~\ref{fig:teasor}, we show how our proposed framework achieves these abilities surpassing conventional ones such as StarGAN~\cite{stargan} and CycleGAN~\cite{CycleGAN2017}.

A first attempt would be to frontalize the pose-variant face input. Despite good visual quality, appearance-based face frontalization methods \cite{drgan,ffgan,tpgan,faceidgan} may suffer from lack of identity-preservation. Geometric modeling methods \cite{hassner15,uvgan2018} faithfully inherit visible appearance but need to guess the invisible appearance due to self-occlusion, leading to extensions like UV-GAN \cite{uvgan2018}.

Further, we note that both texture completion and attribute generation are correlated with 3D shape, that is, the hallucinated appearance should be within the shape area and the generated attribute should comply with the shape. This motivates our framework that utilizes both 3D shape and texture, distinguishing our work from traditional ones that deal only with appearance or UV-GAN that only uses the texture map. 

Specifically, we propose to disentangle the task into mainly two stages: (1) We apply an off-the-shelf 3D shape regression PRNet~\cite{prnet} with a rendering layer to directly achieve 3D shape and weak perspective matrix from a single input, and utilize the information to render partial (self-occluded) texture. (2) A two-step GAN, consisting of a texture completion GAN (TC-GAN) that utilizes the above 3D shape and partial texture to complete the texture map and a 3D Attribute generation GAN (3DA-GAN) that generates target attributes on the completed 3D texture representation. In stage (1), we apply the UV representation~\cite{densereg,prnet} for both 3D point cloud and texture, termed $\mathbf{U}_{p}$ and $\mathbf{U}_{t}$, respectively. The UV representation not only provides the dense shape information but also builds a one-to-one correspondence from point cloud to texture. 

In stage (2), TC-GAN and 3DA-GAN use both $\mathbf{U}_{p}$ and $\mathbf{U}_{t}$ as input to inject 3D shape insights into both the completed texture and generated attribute. Extensive experiments show the effectiveness of our method, which generates geometrically accurate and photorealistic attributes under large pose variation, while preserving identity. 

Our contributions are summarized as the following:
\vspace{-2mm}
\begin{tight_itemize}
\item{We are the first to achieve 3D facial attributes generation under unconstrained head poses such as profile pose. Our method works on the pose-invariant 3D UV space, while most prior ones work on 2D image space.}
\item{We propose a novel two-stage GAN, for UV space texture completion (TC-GAN) and texture attribute generation (3DA-GAN). The stacked structure effectively solves the pose variation problem, conducts face frontalization and can generate attributes for different pose angles.}

\item{We propose a two-phase training protocol to guide the network to focus only on the area related to the attribute, which significantly improves identity-preservation.}
\item{Extensive experiments on several public benchmarks demonstrate the consistently better results in face frontalization, accurate attribute generation, image visual quality and close-to-original identity preservation.}

\end{tight_itemize}

\section{Related Work}
\noindent\textbf{Face Frontalization:} Early works~\cite{hassner15,ferrari2016} apply a 3D Morphable Model and search for dense point correspondence to complete the invisible face region. \cite{xiangyu2015} proposes a high fidelity pose and expression normalization approach based on 3DMM. Sagonas et al.~\cite{sagonas2015} formulate the frontalization as a low rank optimization problem. Yang et al.~\cite{jimei2015} formulate the frontalization as a recurrent object rotation problem. Yim et al.~\cite{yim2015} propose a concatenate network structure to rotate faces with image-level reconstruction constraint. Cole et al.~\cite{cole2017} proposes using the identity perception feature to reconstruct normalized faces. Recently, GAN-based generative models~\cite{drgan,ffgan,tpgan,faceidgan,opensynthesis,uvgan2018} have achieved high visual quality and preserve identity with large extent. Our method aligns in the GAN-based methods but works on 3D UV position and texture other than the 2D images.

\noindent\textbf{Attribute Generation:} Pixel-level graphical editing takes large part in attribute generation. However, we focus on the holistic image-level attribute generation and thus only discuss the closely related works. Li et al.~\cite{cnai} apply an attribute perception loss to guide the attribute synthesis. Upchurch et al.~\cite{upchurch2017} propose the target attribute guided feature-level interpolation for the synthesis. Shen and Liu~\cite{shen2017} introduce residual maps to add or remove specific attributes. GAN-based methods~\cite{icgan,genegan,dnagan,attgan,stargan,fadernet,ganimation,zhang2018generative,xiao2018elegant} aim at connecting the latent attribute code space and the with-target-attribute image space, \ie, swap attribute related latent code~\cite{genegan,dnagan}, or disentangling the attribute for invariant representation~\cite{fadernet}, or imposing an attention network to guide the attribute generation in a specific area~\cite{zhang2018generative}. Xiao et al,~\cite{xiao2018elegant} worked on paired images of attribute transfer. Given low resolution or occluded face images, both \cite{lu2018attribute} and \cite{chen2018high} attempted to generate high resolution images, which satisfy the user-given attributes. Our work lies in the GAN-based methods. In literature, there is no work synthesize attributes based on 3D representation while ours is the first. Moreover, our newly proposed two phase training and masked reconstruction loss, enable the network to focus only on the attribute related region, thus highly preserves the identity.

\begin{figure}[t]
\centering
\includegraphics[width=0.47\textwidth]{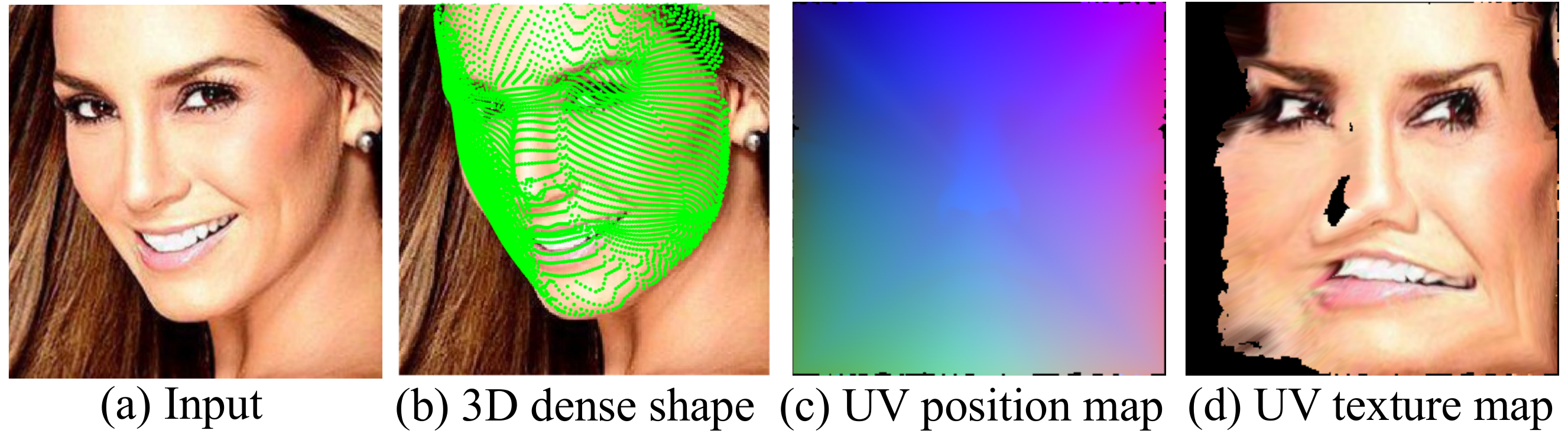}
\caption{Illustration of image coordinate space and UV space. (a) Input image. (b) 3D dense point cloud. (c) UV position map $\mathbf{U}_{p}$ transferred from 3D point cloud. (d) UV texture map $\mathbf{U}_{t}$, partially visible due to pose variation (Best viewed in color).}
\vspace{-5mm}
\label{fig:uv_intro}
\end{figure}

\begin{figure*}[t]
\centering
\includegraphics[width=0.94\textwidth]{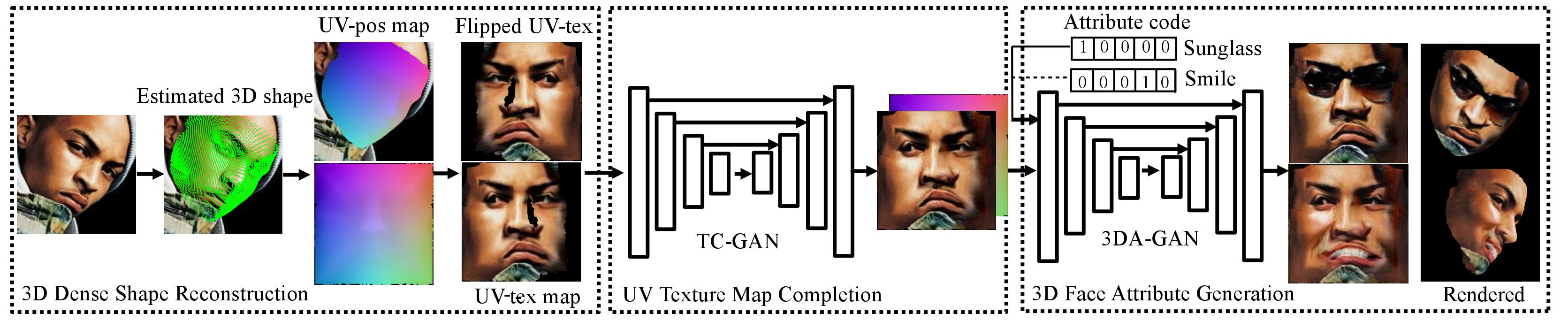}
\caption{The proposed framework of pose-variant 3D facial attribute generation. By 3D dense shape reconstruction, a pose-variant face input is transformed into the UV position map and incomplete UV texture map (with the black holes) due to self-occlusion. Then, a texture completion GAN (TC-GAN) inpaints the black holes into a completed UV texture map. Further, a 3D attribute generation GAN (3DA-GAN) is designed to generate the target attributes on UV texture map and rendered back to 2D images with variant head poses.}
\vspace{-4mm}
\label{fig:flowchart}
\end{figure*}
\section{The Proposed Approach}
In this section, we firstly introduce a dense 3D representation named UV space that supports appearance generation. Then, rendering is conducted to generate visible appearance from the original input. Further, a texture completion GAN is presented to achieve fully visible texture map. In the end, a 3D attribute generation GAN is proposed to work on the 3D UV position and texture representation, generating target attribute under pose-variant conditions.

\subsection{UV Position and Texture Maps}\label{sec:pos_regress}
To faithfully render the visible appearance, we seek a dense 3D reconstruction of shape and texture. The 3D Morphable Model~\cite{Blanz999} sets up a parametric representation by decomposing both shape and texture into linear subspaces. It reduces the space dimension but also drops the high frequency information which is highly demanded for the rendering and generation tasks. Directly applying the raw shape and texture is computationally heavy. Following \cite{densereg,prnet}, we introduce a sphere UV space that homographically map to the coordinate space.

Assume 3D point cloud $\textbf{S}\in\mathbb{R}^{Nx3}$, N is the number of vertices. Each vertex $\textbf{s}=(x,y,z)$ consists of the three dimensional coordinates in 3D space. $(\textbf{u}, \textbf{v})$ are defined as:
\vspace{-2mm}
\begin{align}
\textbf{u}=\frac{1}{\pi}arccos(\frac{x}{\sqrt{x^2 + z^2}}), \textbf{v} = 1 - \frac{1}{\pi}arccos(y)
\label{eq:uv}
\vspace{-2mm}
\end{align}
Eq.~\ref{eq:uv} establishes a unique mapping from dense point cloud to the UV maps. By quantizing the UV space with different granularity, one can control the density of UV space versus the image resolution. In this work, we quantize the UV maps into $256\times256$ and thus preserves $65k$ vertices. As shown in Fig.~\ref{fig:uv_intro}, a UV position map $\mathbf{U}_{p}$ is defined on the UV space, each entry is the corresponding three dimensional coordinate $(x,y,z)$. We apply PRNet~\cite{prnet} to estimate the 3D shape and then exploit Eq.~\ref{eq:uv} to obtain the $\mathbf{U}_p$. A UV texture map $\mathbf{U}_{t}$ is also defined on the UV space, each entry is the corresponding coordinate's RGB color.

\textbf{UV texture map rendering:} $\mathbf{U}_{t}$ of a pose-variant face is partially visible as shown in Fig.~\ref{fig:uv_intro} (d). The invisible region corresponds to the self-occluded region resulting from pose variation. In the original coordinate space, we conduct a z-buffering algorithm~\cite{3ddfa} to label the visible condition of each 3D vertex. Those vertices with largest depth information are  visible while all others are invisible. Assume the visibility matrix $\mathbf{M}$ with entry $1$ means visible and $0$ invisible.

The rendering is a look-up operation by associating the specific coordinate's color to the corresponding $UV$ coordinate. We formulate the process in Eq.~\ref{eq:render}.
\vspace{-2mm}
\begin{align}
\mathbf{U}_{t}(\mathbf{u}, \mathbf{v}) = \mathbf{I}(x,y) \odot \mathbf{M}(x,y,z)
\label{eq:render}
\vspace{-2mm}
\end{align}
where $(\mathbf{u},\mathbf{v})$ is determined by Eq.~\ref{eq:uv} and $\odot$ denotes element-wise multiplication.

\subsection{UV Texture Map Completion} \label{sec:tex_complete}
The incomplete $\mathbf{U}_{t}$ from the rendering is insufficient to conduct the attribute generation. We seek a texture completion that can not only recover photo-realistic appearance but also preserve identity. UV-GAN~\cite{uvgan2018} proposes a similar framework to complete the UV texture map by applying an adversarial network. However, it only considers the texture information. We argue that for 3D UV representation, completing the appearance should consider both texture information and the shape information. For example, combining the original and flipped input will provide a good initialization for appearance prediction. But it only applies the symmetry constraint on shape, which is not sufficient to preserve the shape information. Thus, we take $\mathbf{U}_{p}$, $\mathbf{U}_{t}$ and flipped $\mathbf{U}_{t}$ as input.

\noindent\textbf{Reconstruction module:}
To prepare the UV texture ground truth, we start with near-frontal face images where all the pixels are visible. Then, we perturb the head pose of this original image with random angle. Note that all the pose variant images share the same frontal ground truth which is the original image. By rendering in Eq.~\ref{eq:render}, we obtain the incomplete texture map for the input. Since ground truth is provided, we propose the supervised reconstruction loss to guide the completion.
\vspace{-2mm}
\begin{align}
\mathcal{L}_{r} = \|G_{tc}(\mathbf{U}_{t}, \tilde{\mathbf{U}_{t}}, \mathbf{U}_{p}) - \mathbf{U}_{t}^{*}\|_1
\label{eq:tcgan_recon}
\vspace{-2mm}
\end{align}
$G_{tc}(.)$ stands for the generator consists of the encoder and decoder. $\mathbf{U}_t$ is the partial texture map, $\tilde{\mathbf{U}_{t}}$ the flipped input and $\mathbf{U}_{t}^{*}$ the complete ground truth of the input. 
Merely rely on reconstruction leads to blurry effect. We introduce the adversarial learning to improve the generation quality.

\noindent\textbf{Discriminator module:} Given the ground truth images $\mathbf{U}_{t}^{*}$ in the positive sample set $\mathbf{R}$ and the generated samples $\hat{\mathbf{U}}_{t} = G_{tc}(\mathbf{U}_{t}, \tilde{\mathbf{U}_{t}}, \mathbf{U}_{p})$ in the negative sample set $\mathbf{F}$, we train a discriminator $D$ with the following objective. 
\vspace{-2mm}
\begin{align}
\mathcal{L}_{D} =-\underset{\mathbf{U}_{t}^{*} \in \mathbf{R}}{\mathbb{E}} \log(D(\mathbf{U}_{t}^{*})) - \underset{\hat{\mathbf{U}}_{t} \in \mathbf{F}} {\mathbb{E}}\log(1-D(\hat{\mathbf{U}_{t}}))
\label{eq:tcgan_d}
\vspace{-2mm}
\end{align}

\noindent\textbf{Generator module:} Following the adversarial training, $G_{tc}$ aims to fool D and thus push the objective to the other direction.
\vspace{-2mm}
\begin{align}
\mathcal{L}_{a}=- \underset{\hat{\mathbf{U}_{t}} \in \mathbf{F}}{\mathbb{E}}\log(D(\hat{\mathbf{U}}_{t}))
\label{eq:tcgan_adv}
\vspace{-2mm}
\end{align}

\noindent\textbf{Smoothness term:} To remove the artifact, we propose to apply the total variation loss to locally constrain the smoothness of the output.
\vspace{-2mm}
\begin{align}
\mathcal{L}_{tv}=\frac{1}{|\mathbf{U}_{t}|}\sum |\nabla G_{tc}(\mathbf{U}_{t}, \tilde{\mathbf{U}_{t}}, \mathbf{U}_{p})|
\label{eq:tcgan_tv}
\vspace{-2mm}
\end{align}
$\nabla G_{tc}(\mathbf{U}_{t}, \tilde{\mathbf{U}_{t}}, \mathbf{U}_{p})$ is the gradient of the output $\hat{\mathbf{U}}_{t}$. $|\mathbf{U}_{t}|$ is the number of entries of the texture map. To preserve identity, it is general to introduce a face recognition engine to guarantee the recognition feature of generated image is close to the ground truth feature. In practice, we find the reconstruction constraint Eq.~\ref{eq:tcgan_recon} is sufficient to preserve the identity. It is because major part of the facial area is visible, which already largely indicates the identity information. By symmetry and reconstruction constraint, the identity is well preserved. Thus, the overall loss for TC-GAN is summarized:
\vspace{-2mm}
\begin{align}
\mathcal{L}_{TC}=\lambda_{r} \mathcal{L}_{r} + \lambda_{a} \mathcal{L}_{a} + \lambda_{tv} \mathcal{L}_{tv}
\label{eq:tcgan_all}
\vspace{-2mm}
\end{align}
Weight balance is empirically set as $\lambda_{r}=1, \lambda_{a}=0.1, \lambda_{tv}=0.05$ respectively.

\subsection{3D Face Attribute Generation}\label{sec:attri_generation}
\begin{figure}[t]
\centering
\includegraphics[width=0.48\textwidth]{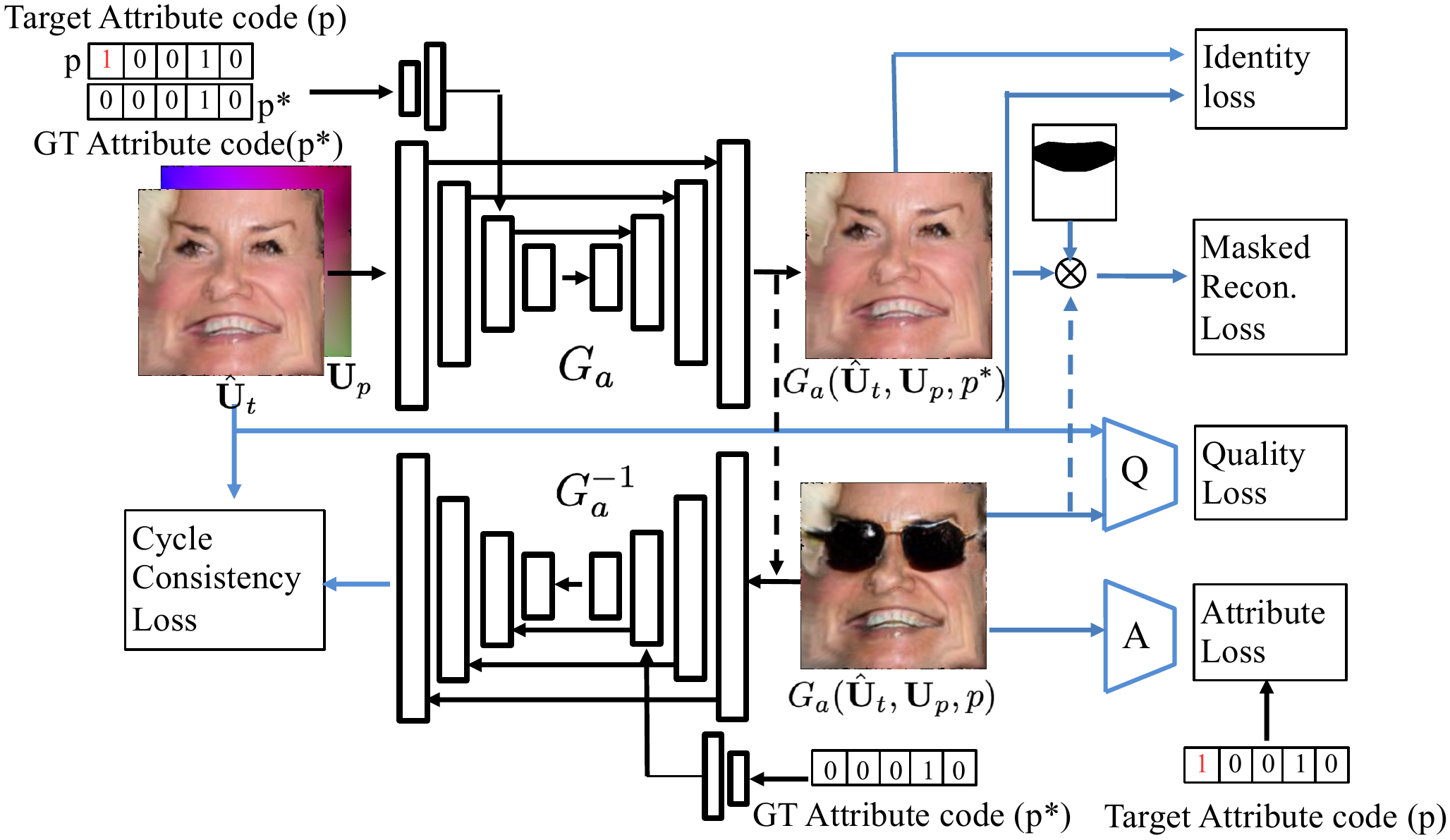}
\caption{The architecture and loss design of 3DA-GAN.}
\vspace{-5mm}
\label{fig:attribute_generation}
\end{figure}

Dissimilar from the traditional image based attribute generation, we adopt the 3D UV representation, the $\mathbf{U}_{p}$ and completed $\hat{\mathbf{U}}_{t}$, as the input. We believe that introducing 3D geometric information can better synthesize attribute, \ie, with 3D shape information, sunglasses will be generated as surface. We formulate the target attribute generation as a conditional GAN framework, as shown in Fig.~\ref{fig:attribute_generation}, by inserting the attribute code $\mathbf{p}$ into the data flow. We manually select 5 out of 40 attributes defined from celebA~\cite{celeba} which do not indicate the face identity. Thus, $\mathbf{p} \in \mathbb{R}^{5}$, each element stands for one attribute, $1$ means with the attribute $0$ without. The attribute code $\mathbf{p}$ is convolved with two blocks and then concatenated to the third block of the encoder of generator $G_{a}$. 

We investigate CycleGAN~\cite{CycleGAN2017} and StarGAN~\cite{stargan} network structures and find that CycleGAN provides a more stable training and better accuracy indicated in experiment section. Thus, we start with the CycleGAN loss design.

\noindent\textbf{Identity loss:} in conditional GAN setting, if input attribute code $\mathbf{p}$ is the original ground truth $\mathbf{p}^{*}$, we expect the output should reconstruct the ground truth input, terming as the identity loss:
\vspace{-2mm}
\begin{align}
    \mathcal{L}_{id} = \|G_{a}(\hat{\mathbf{U}}_{t}, \mathbf{U}_{p}, \mathbf{p}^{*}) - \mathbf{U}_{t}^{*}\|_1
    \label{eq:idloss}
    \vspace{-2mm}
\end{align}

\noindent\textbf{Quality Discriminator:} We introduce a quality discriminator $Q$ in charge of the image quality, leaving the attribute generation correctness to an independent discriminator. The positive sample set $\mathbf{R}_{g}$ are the ground truth $\mathbf{U}_{t}^{*}$ and the negative sample set $\mathbf{F}_{g}$ are the generated UV maps $\mathbf{U}_{g} = G_{a}(\hat{\mathbf{U}}_{t}, \mathbf{U}_{p}, \mathbf{p})$. To update $Q$, we apply the following loss.
\vspace{-2mm}
\begin{align}
\mathcal{L}_{Q} = -\underset{\mathbf{U}_{t}^{*} \in \mathbf{R}_{g}} {\mathbb{E}} \log(Q(\mathbf{U}^{*})) - \underset{\mathbf{U}_{g} \in \mathbf{F}_{g}}{\mathbb{E}}\log(1-Q(\mathbf{U}_{g})))
\label{eq:3dagan_qualityd}
\vspace{-2mm}
\end{align}
The quality loss from $Q$ is fed back to the generator $G_{a}$, resulting the adversarial loss of quality.
\vspace{-2mm}
\begin{align}
\mathcal{L}_{Qa}=- \underset{\mathbf{U}_{g} \in \mathbf{F}_{g}}{\mathbb{E}}\log(Q(\mathbf{U}_{g}))
\label{eq:3dagan_qualityadv}
\vspace{-2mm}
\end{align}

\noindent\textbf{Cycle Consistency:} Following CycleGAN's setting, we simultaneously set an inverse generation module $G_{a}^{-1}$, to convert the generated $\mathbf{U}_{g}$ into the original input $\hat{\mathbf{U}_{t}}$, and expect the converted back UV texture is similar to the original input.
\vspace{-2mm}
\begin{align}
    \mathcal{L}_{cc}=\|G_{a}^{-1}(G_{a}(\hat{\mathbf{U}}_{t}, \mathbf{U}_{p}, \mathbf{p})) - \hat{\mathbf{U}}_{t}\|_1
    \label{eq:cycle_consistency}
    \vspace{-2mm}
\end{align}
Besides the CycleGAN losses, we propose two new losses that specifically deal with attribute generation.

\begin{figure}[t]
\centering
\includegraphics[width=0.48\textwidth]{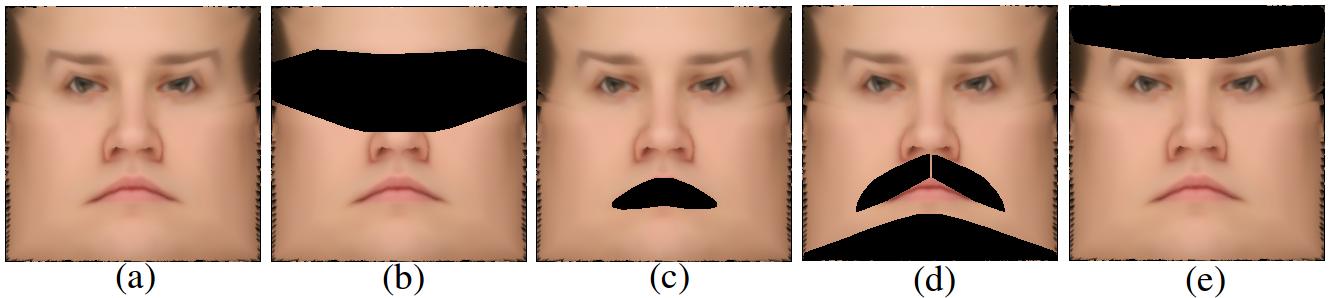}
\caption{Manually defined attribute related masks based on the reference UV texture map. (a) reference $\mathbf{U}_{t}$ (constructed by our generated UV position map and the mean face texture provided by Basel Face Model), (b) eyeglasses mask, (c) lipstick and smile mask, (d) 5'o clock shadow mask, and (e) bangs mask.}
\vspace{-5mm}
\label{fig:mask_template}
\end{figure}
\noindent\textbf{Masked Reconstruction Module:} We manually define the non-attribute area shown in Fig.~\ref{fig:mask_template} on the reference UV texture map. Those $5$ attributes are divided into several different mask types or their combination, \ie, lipsticks and smile share the same mask of Fig.~\ref{fig:mask_template} (c). Together with the fully visible one (mask of all entries as $1$), we define mask $\Omega_{i}, i=1,2,..,5$ indicating all the categories. The reconstruction objective is as below.
\vspace{-2mm}
\begin{align}
\mathcal{L}_{ra} = \|(G_{a}(\hat{\mathbf{U}}_{t}, \mathbf{U}_{p}, \mathbf{p}) - \hat{\mathbf{U}}_{t})\odot \Omega_{i}\|_1, 
\label{eq:3dagan_recon}
\vspace{-2mm}
\end{align}
$i$ is determined by the target attribute code.

\noindent\textbf{Target Attribute Discriminator:} Separated from $Q$, we set an independent discriminator $A$ to evaluate whether the one-bit specific attribute is correctly generated or not. The positive sample set $\mathbf{R}_{a}$ consists of samples from the ground truth with the specific attribute. The negative sample set $\mathbf{F}_{a}$ are the samples generated from $G_{a}$. The target attribute discriminator is updated as:
\vspace{-2mm}
\begin{align}
\mathcal{L}_{A} = -\underset{\mathbf{U}^{*} \in \mathbf{R}_{a}}{\mathbb{E}} \log(A(\mathbf{U}^{*})) - \underset{\mathbf{U} \in \mathbf{F}_{a}}{\mathbb{E}}\log(1-A(\mathbf{U})))
\label{eq:3dagan_attrd}
\vspace{-2mm}
\end{align}
Accordingly, the adversarial loss to update the generator is:
\vspace{-4mm}
\begin{align}
\mathcal{L}_{Av}=- \underset{\mathbf{U} \in \mathbf{F}_{a}}{\mathbb{E}}\log(A(\mathbf{U}))
\label{eq:3dagan_attradv}
\vspace{-4mm}
\end{align}

In TC-GAN, we find that reconstruction loss other than recognition perception loss is sufficient to preserve identity. It also applies for attribute generation. As shown in Fig.~\ref{fig:mask_template}, attribute related area is small portion of the entire facial area. By reconstruction, the large portion already strongly indicates the identity. The overall training is divided in two phases. Phase one accepts the original attribute code and expect to output the reconstructed UV texture. Phase two accepts the target attribute code and generate the image with target attribute.
\vspace{-2mm}
\begin{align}
\mathcal{L}_{p1}=\eta_{id}\mathcal{L}_{id} + \eta_{Qa}\mathcal{L}_{Qa} +  \eta_{cc}\mathcal{L}_{cc} + \eta_{ra}&L_{ra} \\
\mathcal{L}_{p2}=\eta_{id}\mathcal{L}_{id} + \eta_{Qa}\mathcal{L}_{Qa} + \eta_{cc}\mathcal{L}_{cc} + \eta_{ra}&L_{ra} + \eta_{Av}L_{Av} \notag
\label{eq:3dagan_all}
\vspace{-2mm}
\end{align}
The hyper-parameters for phase one and phase two are set as $\eta_{id}=5,\eta_{Qa}=1, \eta_{cc}=10, \eta_{ra}=5$, and $\eta_{id}=5, \eta_{Qa}=1, \eta_{cc}=10, \eta_{ra}=5, \eta_{Av}=1$ respectively.

\section{Implementation Details}
To prepare TC-GAN training data, we collect near-frontal images from 4DFE and 300W-LP (58848 from 4DFE and 2735 from 300W-LP) and augment them with uniformly distributed poses, \ie, from left profile to right profile in every $15^{\circ}$. The near-frontal images are converted to UV representation and serve as the ground truth. The augmented pose-variant images are converted to UV position and incomplete texture, serving as input. By mixing the two training sets, the model generalization ability is enhanced. We apply an hour-glass~\cite{hourglass} structure as the TC-GAN backbone. For structure detail please refer to supplementary material.

We find that inside the structure, skip links are important to preserve high frequency information, especially from the lower layers. 

we train the network using Adam optimizer, with batch size 120 and initial learning rate $1e^{-4}$. It converges within 10 epochs. We further fine-tune it on CelebA training set with initial learning rate $1e^{-5}$ for another 8 epochs.

\begin{figure}[t]
\centering
\includegraphics[width=0.47\textwidth]{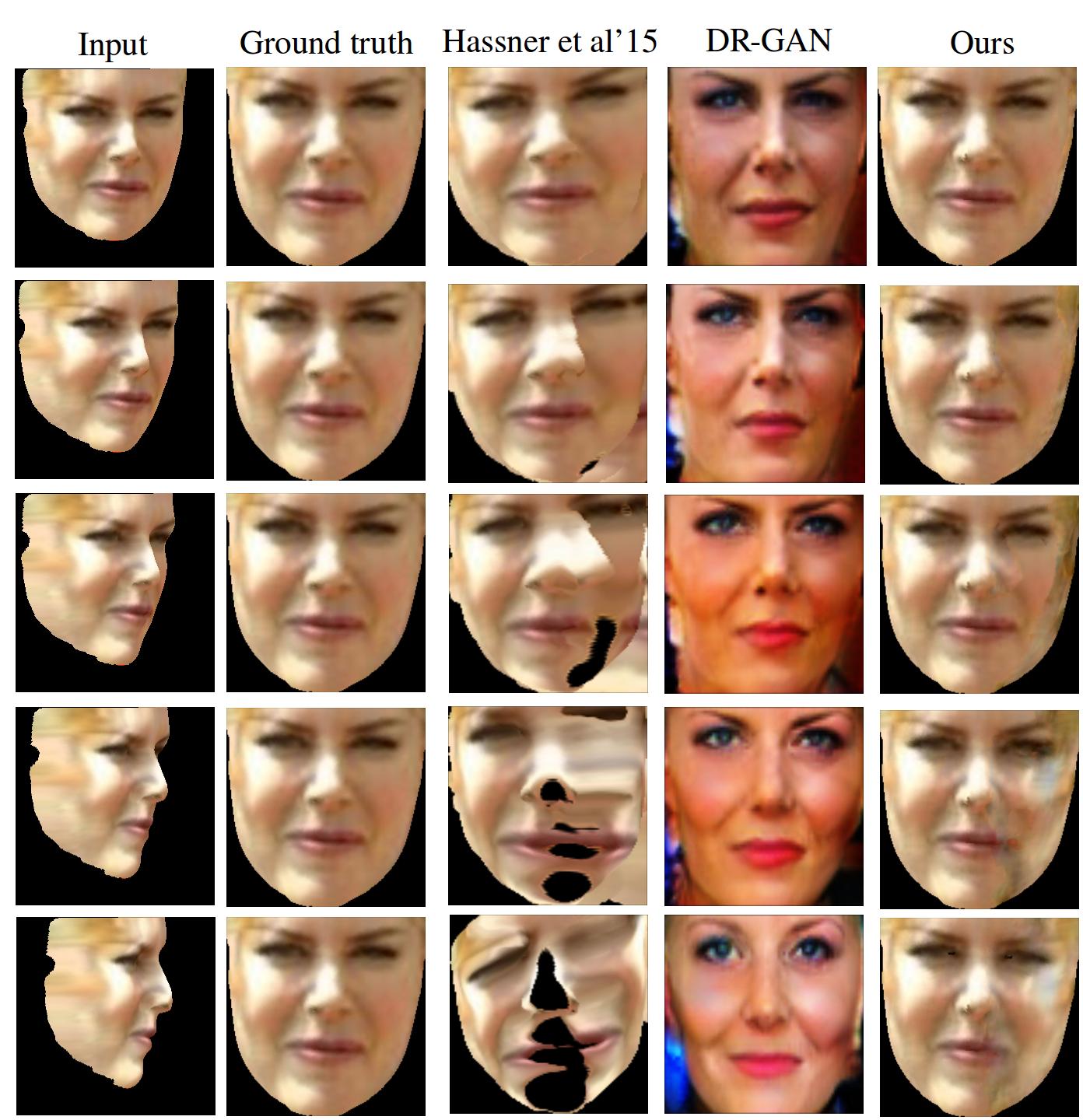}
\caption{Visualization of TC-GAN and other face frontalization methods on LFW~\cite{LFWTech}. A near-frontal image is randomly selected from LFW and shown as ``Ground truth''. We render the ground truth with multiple head poses as input with black background.}
\vspace{-2mm}
\label{fig:texture_visual_lfw}
\end{figure}

\begin{table}[t]
\begin{center}
\small
\begin{tabular}{l|@{\hskip 1mm}c@{\hskip 1mm}|@{\hskip 1mm}c@{\hskip 1mm}|@{\hskip 1mm}c@{\hskip 1mm}|@{\hskip 1mm}c@{\hskip 1mm}|@{\hskip 1mm}c@{\hskip 1mm}} \hline
method & yaw-15 & yaw-30 & yaw-45 & yaw-60 & yaw-75 \\ \hline
Hassner et al.\cite{hassner15} & 30.85 & 53.80 & 174.12 & 208.79 & 203.71 \\
DR-GAN~\cite{drgan} & 82.39 & 84.88 & 90.82 & 98.68 & 110.11 \\
Ours & \textbf{8.06} & \textbf{13.17} & \textbf{20.29} & \textbf{27.39} & \textbf{38.92} \\ \hline
\end{tabular}
\end{center}
\caption{FID score comparison on LFW dataset. We randomly select one image out of the verification pairs, and render yaw to 15, 30, 45, 60, and 75 respectively. FID is calculated between the frontalized images and the not selected original images.}
\vspace{-2mm}
\label{tab:tcgan_res3}
\vspace{-2mm}
\end{table}

We similarly prepare training data for 3DA-GAN training, picking 48K near-frontal images from CelebA for each attribute and convert them to UV representation. Those without target attribute ones serve as input. Those with target attribute ones are positive samples while generated UV texture maps are negative samples for attribute discriminator. For quality discriminator, real UV texture are positive samples and generated ones are negative samples. We randomly select one bit as our target attribute and all others remain not perturbed. The training procedure is two phases: (1) Reconstruction. Assuming input $\hat{\mathbf{U}_{t}}, \mathbf{U}_{p}$ and the original attribute code $\mathbf{p}$. (2) Attribute perturbed generation. We set one attribute per time of $\mathbf{p}$ to be 1. The inputs are $\hat{\mathbf{U}_{t}}, \mathbf{U}_{p}$ and the perturbed $\mathbf{p}^{\prime}$. The two-phase training pushes the generation to focus on the attribute related area while remain the non-attribute area. 

We use Adam optimizer with batch size 16 and initial learning rate $1e^{-4}$. The training converges around 15 epochs across different target attributes. 

\begin{table*}[t]
\begin{center}
\small
\begin{tabular}{c @{\hskip 1.2mm}|@{\hskip 1.2mm} c @{\hskip 1.2mm}| c@{\hskip 1.2mm}c@{\hskip 1.2mm}c@{\hskip 1.2mm}c |c@{\hskip 1.2mm}c@{\hskip 1.2mm}c@{\hskip 1.2mm}c||c@{\hskip 1.2mm}c@{\hskip 1.2mm}c@{\hskip 1.2mm}c | c@{\hskip 1.2mm}c@{\hskip 1.2mm}c@{\hskip 1.2mm}c}
\hline
\multicolumn{2}{c|}{\multirow{2}{*}{Test$\to$}} & \multicolumn{8}{|c||}{F1-score (higher better)} & \multicolumn{8}{c}{FID-score (lower better)}\\ \cline{3-18}
\multicolumn{2}{c|}{} &  \multicolumn{4}{c|}{real (yaw $<$ 45)} & \multicolumn{4}{c||}{real-a. (yaw $\geq$ 45)} & \multicolumn{4}{c|}{real (yaw $<$ 45)} & \multicolumn{4}{c}{real-a. (yaw $\geq$ 45)} \\ \hline 
Model & Train  & SG & LS & SM & BA & SG & LS & SM & BA & SG & LS & SM & BA & SG & LS & SM & BA \\ \hline  
\multirow{1}{*}{FaderNet~\cite{fadernet}} & real & 98.97 & - & - & - & 96.72 & - & - & - & 52.1 & - & - & - & 79.9 & - & - & - \\ 
\multirow{1}{*}{AttGAN~\cite{attgan}} & real & 97.80 & - & - & 86.86 & 91.89 & - & - & 86.15 & 87.6 & - & - & 135.5 & 99.0 & - & - & 172.6 \\ \hline
\multirow{3}{*}{StarGAN~\cite{stargan}*} & real & 97.15 & 84.26 & 87.40 & 89.56 & 96.38 & 77.54 & 77.11 & 86.33 & 85.7 & 78.9 & 92.3 & 82.3 & 139.8 & 135.9 & 150.6 & 144.0\\ 
& real-a & 97.35 & 78.87 & 83.40  & 89.33 & 98.07 & 75.43 & 79.01 & 86.77 & 72.7 & 68.9 & 58.9 & 59.5 & 114.0 & 85.1 & 82.8 & 105.3 \\
& Ours & \textbf{98.88} & \textbf{84.70} & \textbf{87.87} & \textbf{94.86} & \textbf{98.23} & \textbf{82.04} & \textbf{83.32} & \textbf{93.67} & \textbf{38.2} & \textbf{34.1} & \textbf{33.0} & \textbf{21.8} & \textbf{36.3} & \textbf{35.4} & \textbf{30.6} & \textbf{19.4} \\ \hline 
\multirow{3}{*}{CycleGAN~\cite{CycleGAN2017}*} & real & 97.66 & 84.41 & 86.33  & 70.96  & 90.49 &  74.45 & 76.48  & 69.01 & 30.1 & 25.1 & 32.3 & 28.7 & 40.9 & 49.2 & 43.3 & 36.8 \\ 
& real-a & 98.93 & 91.34 & 84.25  & 82.43  & 97.31 & 69.27 & 75.51 &  80.70 & 33.9 & \textbf{12.5} & \textbf{12.7} & \textbf{9.1} & \textbf{19.8} & 31.0 & 17.1 & 11.5 \\
(ResNet) & Ours & \textbf{99.37} & \textbf{94.69} & \textbf{94.56}  & \textbf{93.35}  & \textbf{99.10} & \textbf{93.04} & \textbf{91.49} &  \textbf{91.64} & \textbf{18.5} & \textbf{12.6} & \textbf{13.0} & \textbf{10.3} & 29.7 & \textbf{10.9} & \textbf{11.0} & \textbf{8.9} \\  \hline 

\end{tabular}
\end{center}
\caption{Quantitative comparison on attribute generation by F1 score and FID~\cite{fid2017} score from CelebA testing set. The target generated attribute is evaluated by an off-line attribute classifier for F1 score (precision and recall). Visual quality is indicated by FID score between the target attribute generated images and the ground truth with same attribute images. ``real'' means original CelebA training set.  ``real-a'' means original plus pose augmented images. ``Ours'' means training with our proposed loss and UV texture data. *: we apply the network structure and re-train models. SG: Sunglass, LS: Wearing Lipstick, SM: Smiling, BA: Bangs.}
\label{tab:attr_real}
\vspace{-3mm}
\end{table*}

\begin{table*}[t]
\begin{center}
\small
\begin{tabular}{@{}c |@{\hskip 1.5mm} c | c@{\hskip 1.5mm}c@{\hskip 1.5mm}c@{\hskip 1.5mm}c |c@{\hskip 1.5mm}c@{\hskip 1.5mm}c@{\hskip 1.5mm}c || c@{\hskip 1.5mm}c@{\hskip 1.5mm}c@{\hskip 1.5mm}c | c@{\hskip 1.5mm}c@{\hskip 1.5mm}c@{\hskip 1.5mm}c}
\hline
\multicolumn{2}{c|}{\multirow{2}{*}{Test$\to$}} & \multicolumn{8}{|c||}{F1-score (higher better)} & \multicolumn{8}{c}{FID-score (lower better)}\\ \cline{3-18}
\multicolumn{2}{c|}{} &  \multicolumn{4}{c|}{ real (yaw $<$ 45) }  & \multicolumn{4}{c||}{ real-a (yaw $\geq$ 45)}  & \multicolumn{4}{c|}{ real (yaw $<$ 45) }  & \multicolumn{4}{c}{ real-a (yaw $\geq$ 45)}\\ \hline 
Model & Loss & SG & LS & SM & BA & SG & LS & SM & BA & SG & LS & SM & BA & SG & LS  & SM & BA\\ \hline  
\multirow{3}{*}{CycleGAN} & w/o 
Eq.~\ref{eq:3dagan_recon}~\ref{eq:3dagan_attradv} & 97.97 & 87.92 & 84.62  & 83.65  & 97.93 &  86.21 & 81.11 & 82.21 & 20.2 & \textbf{10.6} & \textbf{7.8} & 14.1  & 43.8 & 20.4 & 27.2 & 18.3 \\
 & w/o Eq.~\ref{eq:3dagan_recon} & 99.28 & 92.95 & 93.17 & \textbf{94.86} & 98.87 & 90.79 & 89.50 & \textbf{93.82} & \textbf{17.6} & 17.5 & 13.8  & 11.9  & \textbf{26.6} & 18.1 & 15.0 & 11.4 \\
(ResNet) & w/o Eq.~\ref{eq:3dagan_attradv} & 97.82 & 83.28 & 81.81 & 86.56 & 97.54 & 82.35 & 78.43 & 85.86 & 29.1 & 19.0 & 18.1 & 10.5 & 39.3 & 18.4 & 17.7 & 10.4 \\
 & Full & \textbf{99.37} & \textbf{94.69} & \textbf{94.56}  & \textbf{93.35}  & \textbf{99.10} & \textbf{93.04} & \textbf{91.49} &  \textbf{91.64} & \textbf{18.5} & \textbf{12.6} & \textbf{13.0} & \textbf{10.3} & \textbf{29.7} & \textbf{10.9} & \textbf{11.0}  & \textbf{8.9}\\ \hline 
\end{tabular}
\end{center}
\caption{Ablation study for w/o masked reconstruction loss (Eq.~\ref{eq:3dagan_recon}), and/or w/o attribute loss (Eq.~\ref{eq:3dagan_attradv}). We put the CycleGAN loss (w/o Eq.~\ref{eq:3dagan_recon}~\ref{eq:3dagan_attradv}) as starting point, \ie quality adversarial loss, identity loss and cycle consistency loss, since we believe the CycleGAN loss ablation is fully studied in~\cite{CycleGAN2017}. F1 and FID scores are reported. We use CycleGAN ResNet structure as it achieves the best result across the experiments. SG: Sunglass, LS: Wearing Lipstick, SM: Smiling, BA: Bangs.} 
\label{tab:ablation}
\vspace{-3mm}
\end{table*}

\begin{table}[h]
\begin{center}
\small
\begin{tabular}{c@{\hskip 1mm}|@{\hskip 1mm}c@{\hskip 1mm}|@{\hskip 1mm}c@{\hskip 1mm}|@{\hskip 1mm}c@{\hskip 1mm}|@{\hskip 1mm}c@{\hskip 1mm}|@{\hskip 1mm}c@{\hskip 1mm}|@{\hskip 1mm}c} \hline
 Method & SG & LS & SH & SM & BA & Avg.  \\ \hline
Original & - & - & - & - & - & 91.38  \\ \hline
FaderNet~\cite{fadernet} & 79.05 & - & - &  - & - & 79.05  \\ 
AttGAN~\cite{attgan} & 87.94  & - & - & - & 82.20 & 85.07  \\
StarGAN~\cite{stargan}* & 75.28 & 78.11 & 81.11 & 78.80 & 81.31 & 79.03 \\
CycleGAN~\cite{CycleGAN2017}* & 89.79 & \textbf{88.09} & 88.48 & 90.00 & 89.20 & 89.11 \\ \hline
Ours & \textbf{90.40} & 87.11 & \textbf{89.76} & \textbf{90.68} & \textbf{90.06} & \textbf{89.60} \\ \hline
\end{tabular}
\end{center}
\caption{Identity preserving evaluation on IJBA dataset under the verification protocol, reporting TAR@FAR0.01. *: models we retrain on our training data. SG: Sunglass, LS: Wearing Lipstick, SH: 5'oclock shadow, SM: Smiling, BA: Bangs.}
\label{tab:idp_ijba}
\vspace{-5mm}
\end{table}

\section{Experiments}
In this section, we evaluate our framework for the tasks of UV texture map completion and the 3D attribute generation. Regarding the training, for texture completion, we generate the UV space representation of 300W-LP~\cite{3ddfa} and 4DFE~\cite{4dfe} to form our training set.

The evaluation for texture completion is conducted on LFW~\cite{LFWTech} on both visualization and FID score as a fair comparison to other methods.  For attribute generation, we generate the UV space representation of CelebA~\cite{celeba} and provide the rendered pose augmented images for both training and testing.

\subsection{Datasets}
\noindent\textbf{300W-LP}: It is generated from 300W~\cite{300w2013} face database by 3DDFA~\cite{3ddfa}, in which it establishes a 3D morphable model and reconstructs the face appearance with varying head poses.

It consists of overall 122,430 images from 3,837 subjects. For each subject, images are with uniformly distributed varying head poses.

\noindent\textbf{CelebA}: It contains about 203K images with 40 attributes per image annotated. 
The distribution of this dataset in terms of yaw angle is highly long-tailed towards near-frontal, which remains the demand to augment it for more pose-variant attribute generation.

\noindent\textbf{4DFE}: It is a high-resolution 3D dynamic facial expression database. 
It contains 606 3D facial expression sequences captured from 101 subjects, with a total of approximately 60,600 frame models. Each 3D model of a 3D video sequence has the resolution of approximately 35,000 vertices. The texture video has a resolution of about $1040\times1329$ pixels per frame.

\subsection{UV Texture Map Completion}
In our framework, we firstly apply a 3D dense shape reconstruction and rendering to obtain a partially visible UV texture map. Then we apply our TC-GAN to obtain the completed UV texture map and render it back to image-level appearance. 

\textbf{Frontalization Visual Comparison:} since our framework provides a way to conduct face frontalization, we visually compare our method with several state-of-the-art frontalization methods in Fig.~\ref{fig:texture_visual_lfw}. The traditional geometric method~\cite{hassner15} fails to complete the holes caused by self-occlusion when head pose is large. DR-GAN~\cite{drgan} works fairly well when head pose is small. When head pose is close to profile, DR-GAN fails to preserve the face identity while our method consistently preserves the identity across different head poses. Our method also consistently preserves the skin color where DR-GAN cannot.

\textbf{Quantitative Comparison:} the Fletcher Inception Distance (FID)~\cite{fid2017} is introduced in Table~\ref{tab:tcgan_res3} to quantitatively indicate the photo-realisticity of generated images compared to original real images. The closer to real images, the lower FID score. In Table~\ref{tab:tcgan_res3}, our method clearly achieves significantly lower FID score than other methods.

\begin{figure*}
\centering
\includegraphics[width=1\textwidth]{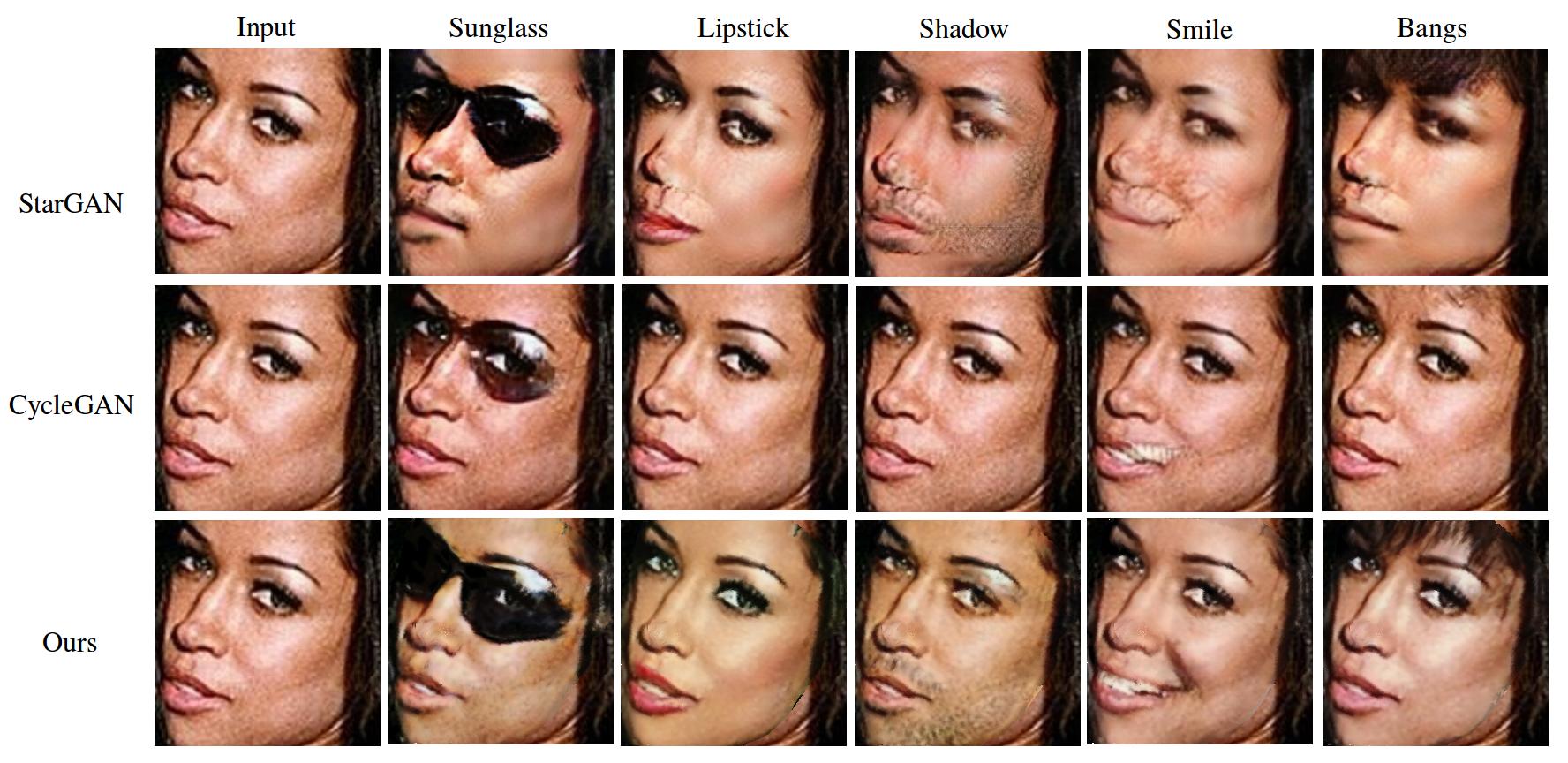}
\caption{Pose-variant qualitative results of our 3DA-GAN compared to StarGAN~\cite{stargan} and CycleGAN~\cite{CycleGAN2017} trained on our prepared data. }
\vspace{-2mm}
\label{fig:attr_visual_1}
\end{figure*}

\begin{figure*}
\centering
\includegraphics[width=1\textwidth]{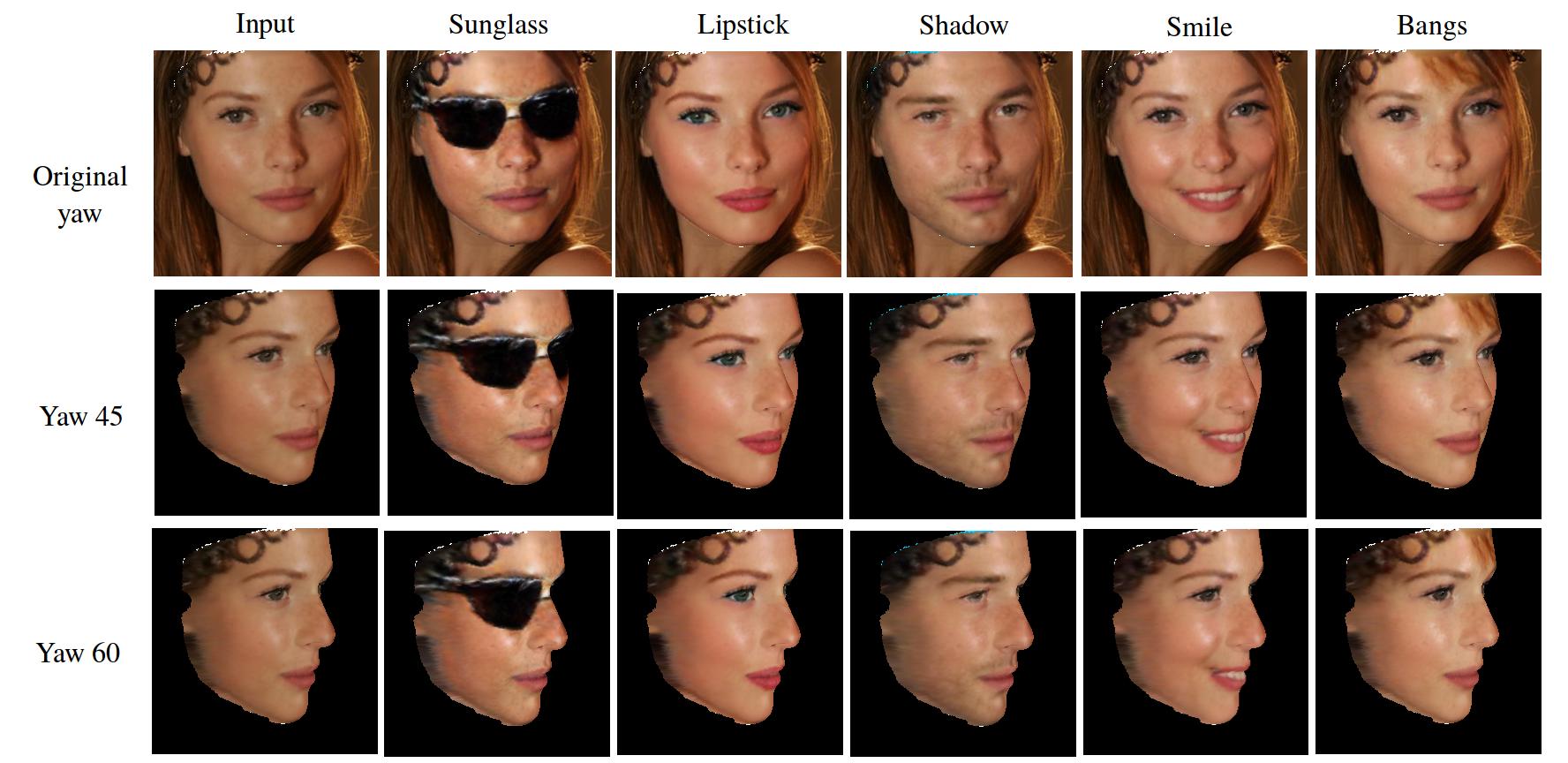}
\caption{Visual results of applying our method to augment face images from CelebA~\cite{celeba} testing set, in attributes and yaw angles.}
\vspace{-2mm}
\label{fig:attr_pose_1}
\end{figure*}

\subsection{3D Attribute Generation}
We manually select $5$ out of $40$ attributes defined in CelebA, which do not indicate face identity and only correlate with the facial area. They are Sunglasses (SG), Wearing lipstick (LS), 5'o clock shadow (SH), Smiling (SM), and Bangs (BA). We strictly follow the CelebA training, validation and testing splitting protocol. \footnote{SH is not shown in Table~\ref{tab:attr_real} and Table~\ref{tab:ablation} due to space limit. Please refer to supplementary material for complete information.}

Traditional attribute generation methods, \ie, FaderNet~\cite{fadernet} and AttGAN~\cite{attgan}, are trained on 2D images. 

For fair comparison, we apply StarGAN and CycleGAN network structures, trained orthogonally on real, real plus pose augmented images and our UV texture and position maps. For real data in CelebA, we observe a strong head pose bias towards near-frontal poses. We calculate the pose by the reconstructed 3D shape vertices, and use it to split the testing data into $yaw < 45^{\circ}$ and $\ge 45^{\circ}$. As $yaw \ge 45^{\circ}$ testing data is very few, such as, $221$ images for lipstick, where $yaw < 45^{\circ}$ have $9288$ images, we augment the $yaw \ge 45^{\circ}$ data from near-frontal images and achieves $6735$ augmented samples to match the volume of $yaw < 45^{\circ}$.

\textbf{Attribute Generation Accuracy:} we apply an off-line attribute classifier, trained on CelebA training set, to evaluate the attribute generation performance, whose average precision on CelebA testing set is $91.7\%$, close to state-of-the-art performance. F1 score is reported as precision and recall may vary due to threshold setting.

We apply 3DA-GAN on the negative samples (without the target attribute) to generate the images with target attribute, serving as positive samples. Further, FID~\cite{fid2017} is computed to evaluate the photo-realisticity of the attribute augmented images. 

In Table~\ref{tab:attr_real}, we compare to several state-of-the-arts, FaderNet~\cite{fadernet}, AttGAN~\cite{attgan}, StarGAN~\cite{stargan} and CycleGAN~\cite{CycleGAN2017}. The last two are retrained on original celebA real data, real plus pose augmented data (``real-a'') as well as our UV texture and position data. For ``Ours'', we apply our proposed loss instead of StarGAN or CycleGAN loss. The numbers in Table~\ref{tab:attr_real} clearly show that our proposed 3DA-GAN consistently achieves higher F1 score than the state-of-the-arts. Moreover, our method also achieves consistently lower FID score. On CycleGAN (ResNet) model, our method FID score is close to the one trained on ``real-a'', \ie tie on $yaw < 45$ and slightly better on $yaw \ge 45$. However, our method achieves much higher F1 score (precision and recall), \ie more than $10\%$ higher on ``SM'' and ``BA'', compared to CyCleGAN trained on ``real-a'' across yaw $< 45$ and $\ge 45$.

\textbf{Identity Preserving Property:} we apply a state-of-the-art face recognition engine, ArcFace~\cite{arcface} to provide the identity feature. For each verification pair, we randomly select one image without the target attribute, apply our method to generate the target attribute, and evaluate the similarity between the generated target attribute image and the not selected image. We independently run experiments for those 5 attributes. In Table~\ref{tab:idp_ijba}, ``Original'' means the original verification accuracy without any attribute generation, which serves as the upper bound for all methods. Compared to other methods, our 3DA-GAN achieves almost all higher verification accuracy while slightly worse on lipstick. Nevertheless, our method achieves $89.60\%$ average accuracy, which is close to the upper bound $91.38\%$, indicating that the proposed attribute generation maximumly preserves identity information.

\textbf{Visualization:} we show a pose-variant face attribute generation example in Fig.~\ref{fig:attr_visual_1}, and compare to StarGAN and CycleGAN. The 2D image based methods suffers from the pose variation, \ie, for both StarGAN and CycleGAN in Sunglass, the left eye region is not correctly generated. In smile, StarGAN failed to generate the attribute while CycleGAN shows unpleasant artifacts in the mouth area. In contrast, our method shows not only the correct attribute generation but also the pleasant visual quality. Worth noting that for ``lipstick'' and ``shadow'', they are actually related to the gender or identity. This is because for lipstick, the dataset is naturally biased towards female. For shadow, the training images are quite similar to another attribute ``beard'', which caused the similar appearance generation effect.

Further shown in Figure~\ref{fig:attr_pose_1}, given an unconstrained face image, our method can generate target attribute with variant head poses. It provides strong potential in high quality face editing of multiple attributes and can serve as face augmentation for face recognition alongside head pose and attribute axis.

\subsection{Ablation Study}
We investigate the contribution of each component proposed in our framework. In Table~\ref{tab:ablation}, we start with the default CycleGAN loss, which is without our proposed masked reconstruction loss Eq.~\ref{eq:3dagan_recon} and attribute adversarial loss Eq.~\ref{eq:3dagan_attradv}. For CycleGAN loss, \ie, generative adversarial loss (a.k.a quality adversarial loss), identity loss and cycle consistency loss, we believe these components' effects are clearly discussed in ~\cite{CycleGAN2017}. Thus, we focus on the two newly proposed losses Eq.~\ref{eq:3dagan_recon} and Eq.~\ref{eq:3dagan_attradv}. Overall, without each or both of the two new components, the performance across F1 and FID score is degraded in certain degree. Moreover, without attribute adversarial loss is more critical as accuracy drops significantly more than without masked reconstruction loss.

\section{Conclusion}
We propose a two-stage Texture Completion GAN (TC-GAN) and 3D Attribute GAN (3DA-GAN), to tackle the pose-variant facial attribute generation problem. The TC-GAN inpaints the missing appearance from self-occlusion and provides a normalized UV texture. Our 3DA-GAN works on the UV texture space to generate target attributes with maximum preserved subject identity. Extensive experiments show that our method achieves consistently better attribute generation accuracy, closer to original images' visual quality, and higher identity preserving verification accuracy, when compared to several state-of-the-art attribute generation methods. Our good generation quality also provides the potential for face editing and face image augmentation alongside pose and attribute axis.

{\small
\bibliographystyle{ieee}

}

\appendix

\section{3D Shape Alignment and UV Maps Rendering}
\begin{figure*}
\centering
\includegraphics[width=0.9\textwidth]{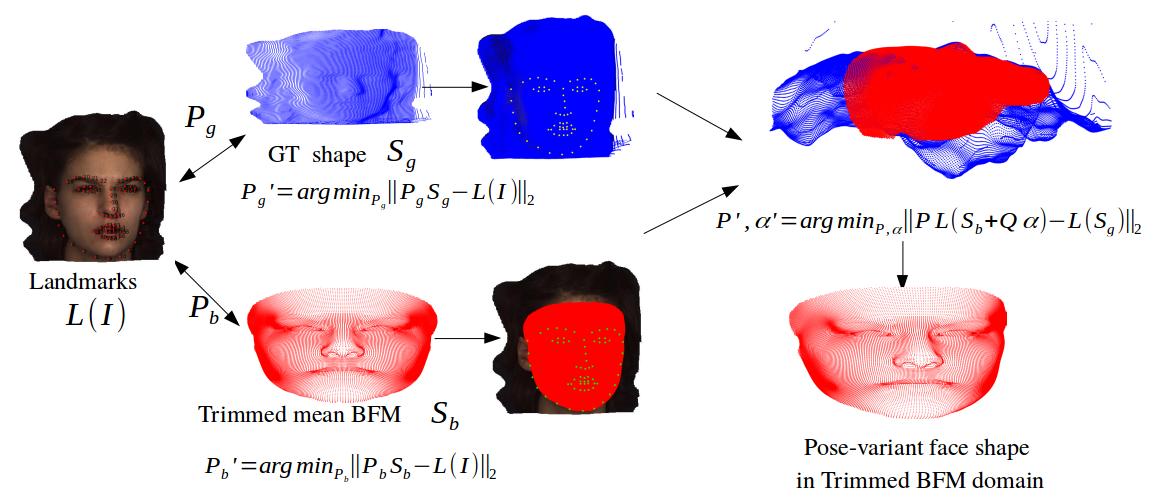}
\caption{Align a ground truth shape or an estimated shape from the existing 3D reconstruction method to the trimmed BFM shape. The example image is from 4DFE dataset and the landmarks $L(I)$ can be obtained by any off-the-shelf image based landmark detector.}
\vspace{-2mm}
\label{fig:point_cloud_align}
\end{figure*}

In this section, we explain how we prepare our ground truth 3D point cloud with respect to the reference BFM 3D shape. We first trim the original BFM shape to the one focusing on the facial area and consists of 38K vertices, as the BFM reference shape thereafter. Given an image, we obtain its 3D shape from the dataset or estimated by \cite{prnet}. Since the number and definition of 3D vertices are different, the untrimmed shape need to be aligned to the reference trimmed BFM shape. A diagram for this alignment is shown in Fig.~\ref{fig:point_cloud_align}. 

The 4DFE 3D point cloud and reference BFM are deformed to match the detected 2D landmarks.
Then we refine the alignment via a 3D-ICP like procedure to obtain the aligned shape.

\begin{figure*}
\centering
\includegraphics[width=0.9\textwidth]{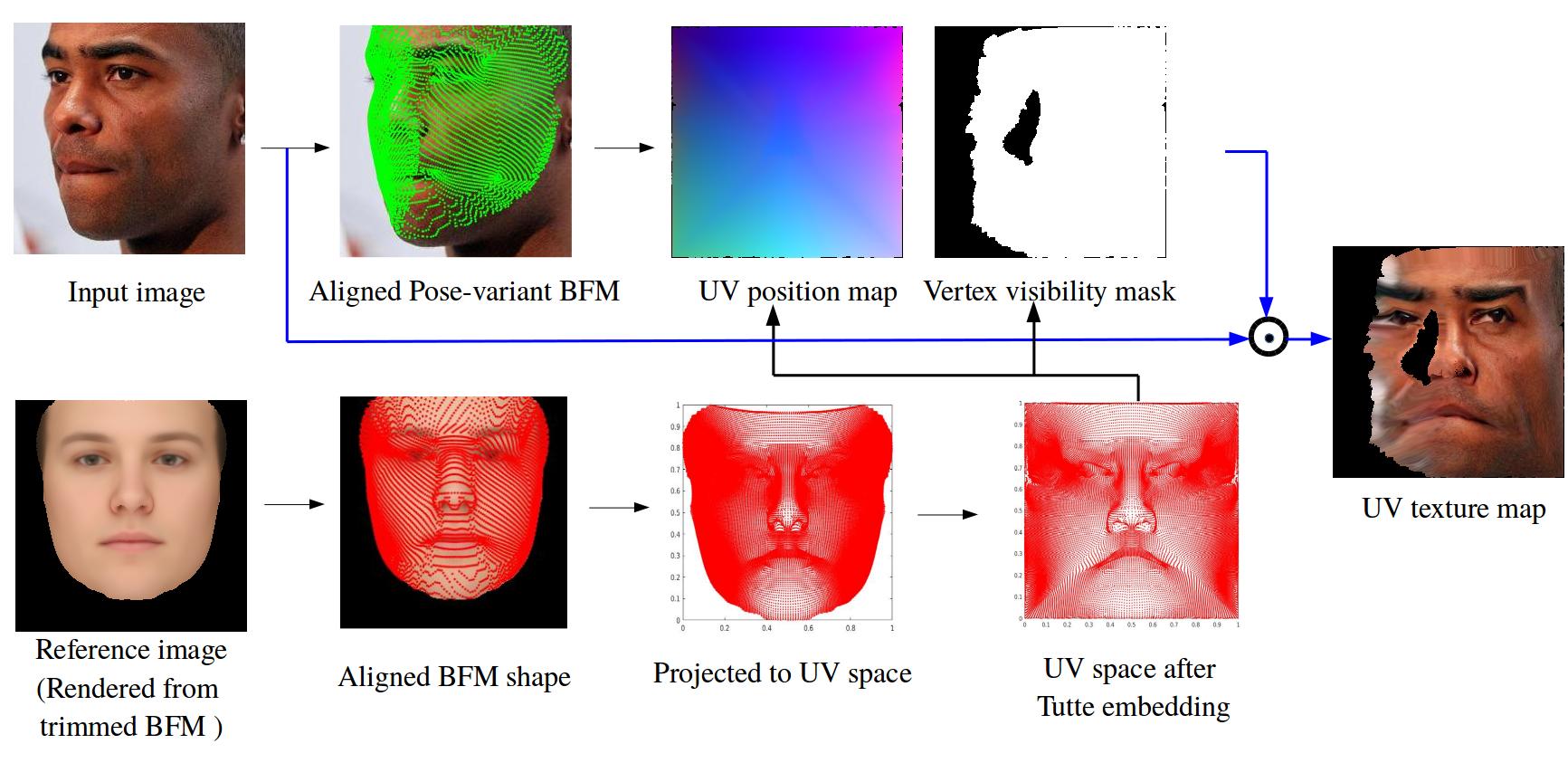}
\caption{Given an input image, conversion from the aligned BFM to the fixed UV coordinates, and the uv texture map rendering based on the vertex visibility and input image. $\odot$ denotes element-wise multiplication.}
\vspace{-2mm}
\label{fig:tutte_render}
\end{figure*}

Given the aligned shape, our goal is to obtain a 3D dense shape representation, i.e., UV texture map, so that the high frequency information can be preserved. To this end, a reference UV coordinates is introduced as illustrated in the lower part of Fig.~\ref{fig:tutte_render}. By extrapolating 3D points based on this reference coordinates and the aligned pose-variant shape, we can get very high resolution UV position map. Here we set it as $256\times256$. Note that this reference UV coordinates is shared by all images, so every pixel corresponds to the same facial point; this is essential to define the attribute related masks (Fig. 5 of the main paper). It enables the attribute generation under an invariant UV space, where arbitrary head pose variation is allowed for the input.

\section{Identity Preserving Evaluation for TC-GAN}

We have already shown in Section 5.2 of the main paper that our TC-GAN can achieve better quality of frontalization with the lowest FID score compared to \cite{hassner15} and DR-GAN~\cite{drgan}. Here, we take a step further to evaluate the verification accuracy on LFW dataset by applying all methods to the non-frontal images, which we define as the ones of yaw $\ge 15$ and replacing the original images with the frontalized ones. Again, the state-of-the-art face recognition engine, ArcFace~\cite{arcface} is exploited to provide the identity features. In Table~\ref{tab:tcgan_ip}, the accuracy based on TC-GAN drops the least compared to original performance, which indicates our method preserves identity better than the state-of-the-arts.

\begin{table}[t]
\begin{center}
\small
\begin{tabular}{l|c} \hline
method & Verification Accuracy \\ \hline
Original & $99.27\pm 0.11$ \\
Hassner et al.\cite{hassner15} & $98.91\pm 0.15$ \\
DR-GAN~\cite{drgan} & $96.43\pm 0.55$ \\
Ours & $\textbf{99.17}\pm 0.12$ \\ \hline
\end{tabular}
\end{center}
\caption{Verification accuracy comparison on LFW dataset. We apply our TC-GAN and other face frontalization methods to the LFW images of yaw angle $\ge 15$ to replace the original image with the frontalized one.}
\vspace{-2mm}
\label{tab:tcgan_ip}
\vspace{-2mm}
\end{table}

\section{More Attribute Generation and Pose-variant Attribute Augmentation Results}

\begin{figure*}
\centering
\includegraphics[width=1\textwidth]{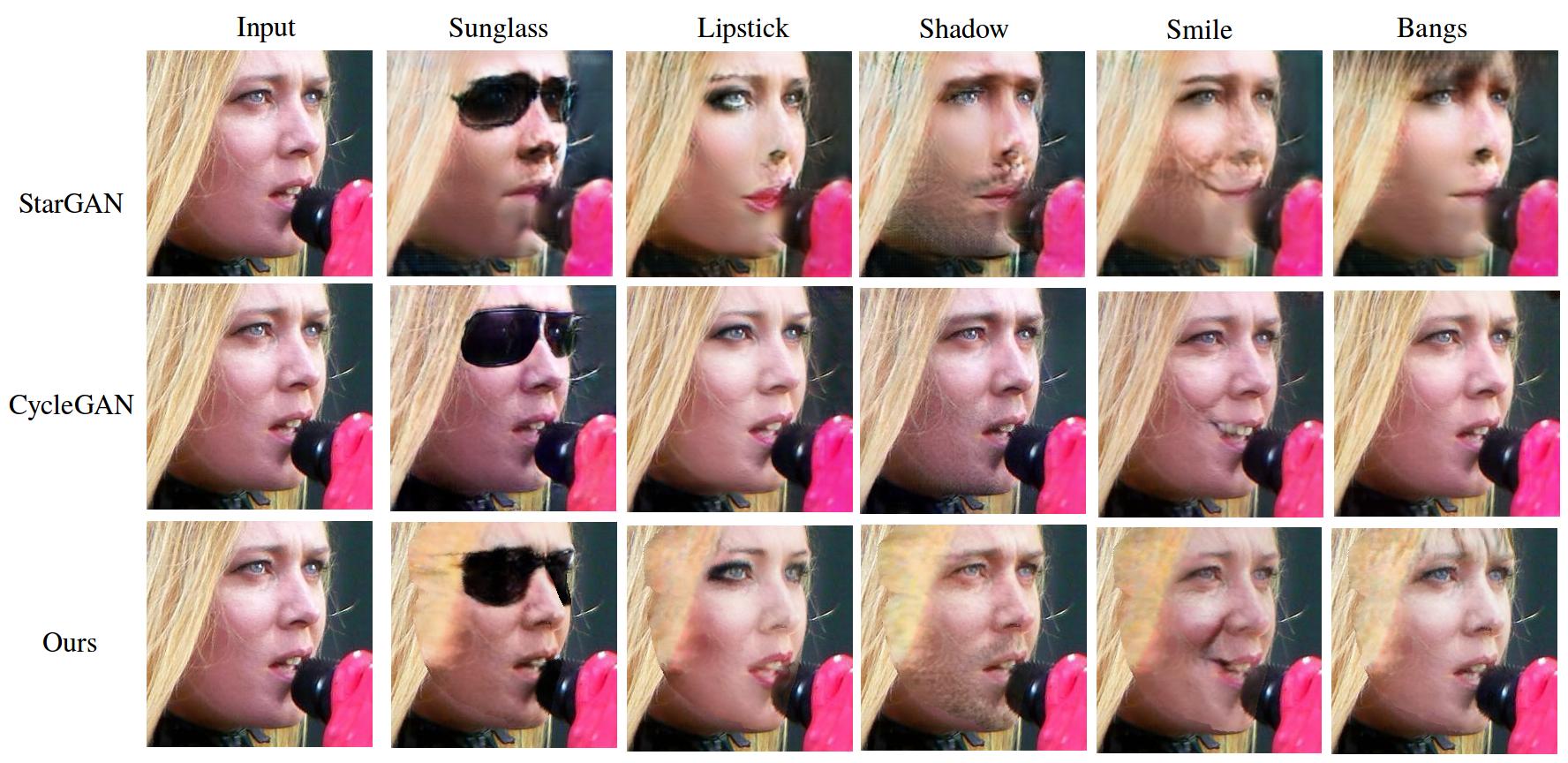}
\caption{Pose-variant qualitative results of our 3DA-GAN compared to StarGAN~\cite{stargan} and CycleGAN~\cite{CycleGAN2017} trained on our prepared data. }
\vspace{-2mm}
\label{fig:attr_visual_2}
\end{figure*}

\begin{figure*}
\centering
\includegraphics[width=1\textwidth]{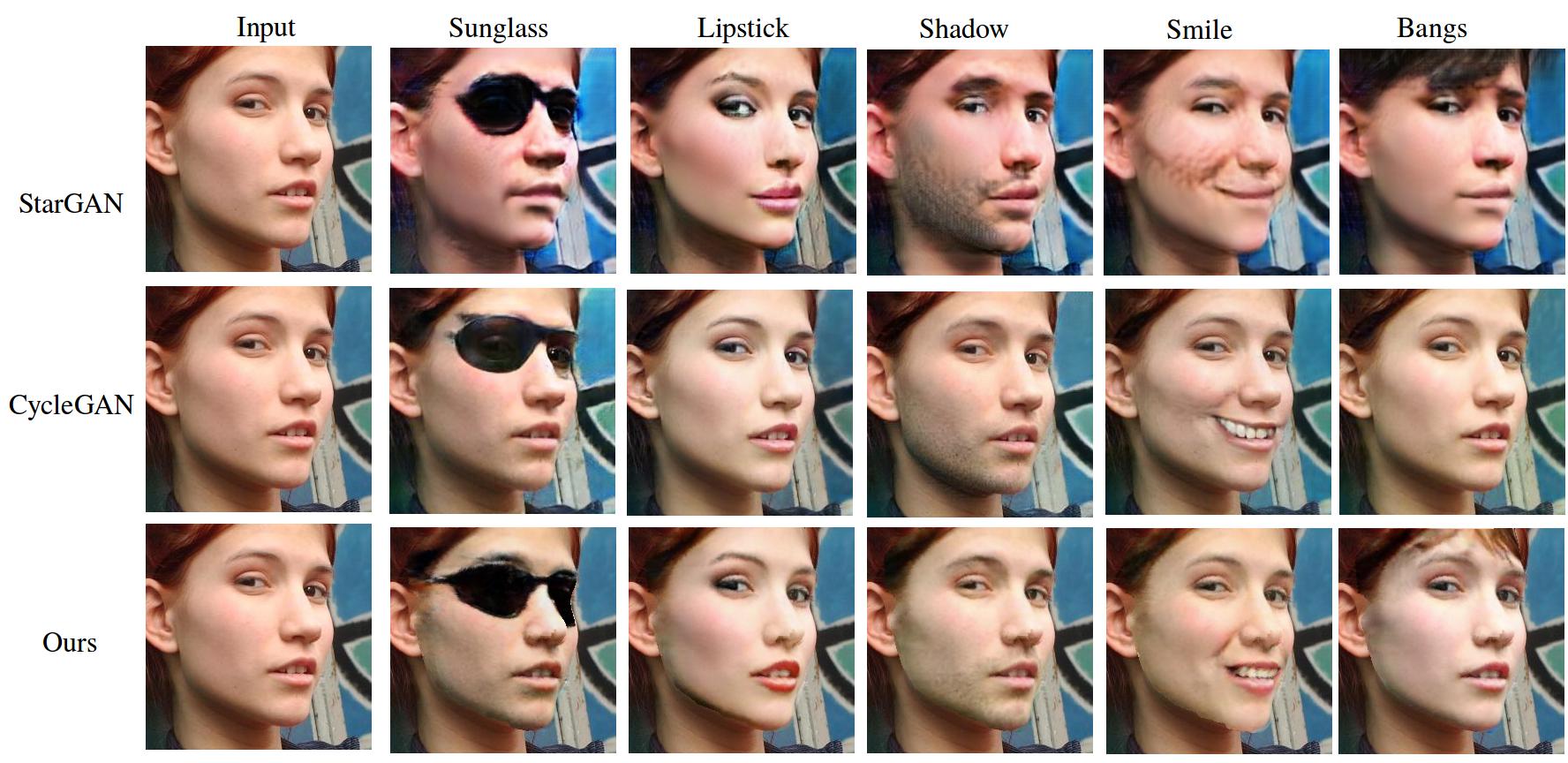}
\caption{Pose-variant qualitative results of our 3DA-GAN compared to StarGAN~\cite{stargan} and CycleGAN~\cite{CycleGAN2017} trained on our prepared data. }
\vspace{-2mm}
\label{fig:attr_visual_3}
\end{figure*}

\begin{figure*}
\centering
\includegraphics[width=1\textwidth]{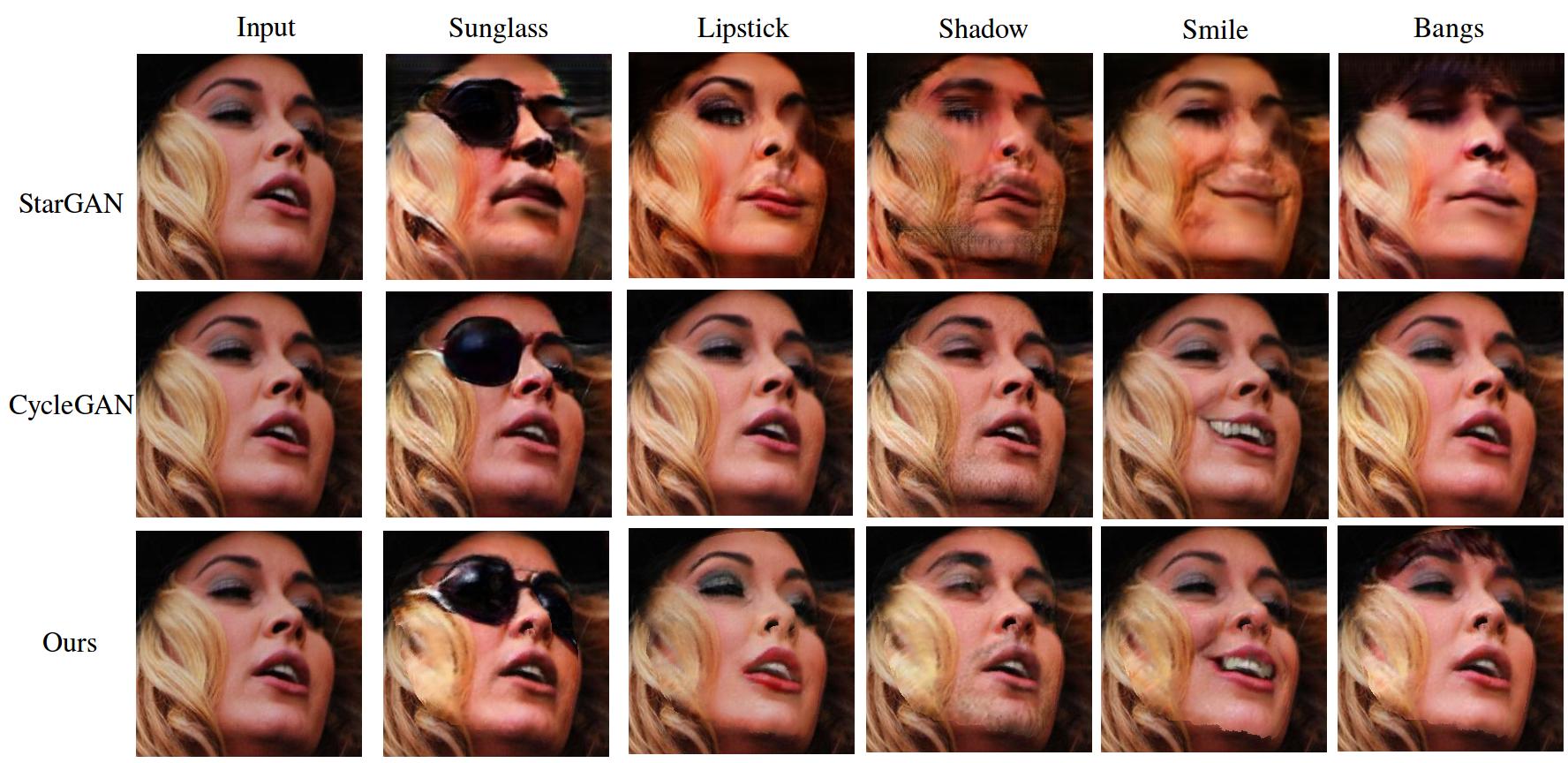}
\caption{Pose-variant qualitative results of our 3DA-GAN compared to StarGAN~\cite{stargan} and CycleGAN~\cite{CycleGAN2017} trained on our prepared data. }
\vspace{-2mm}
\label{fig:attr_visual_4}
\end{figure*}

\begin{figure*}
\centering
\includegraphics[width=1\textwidth]{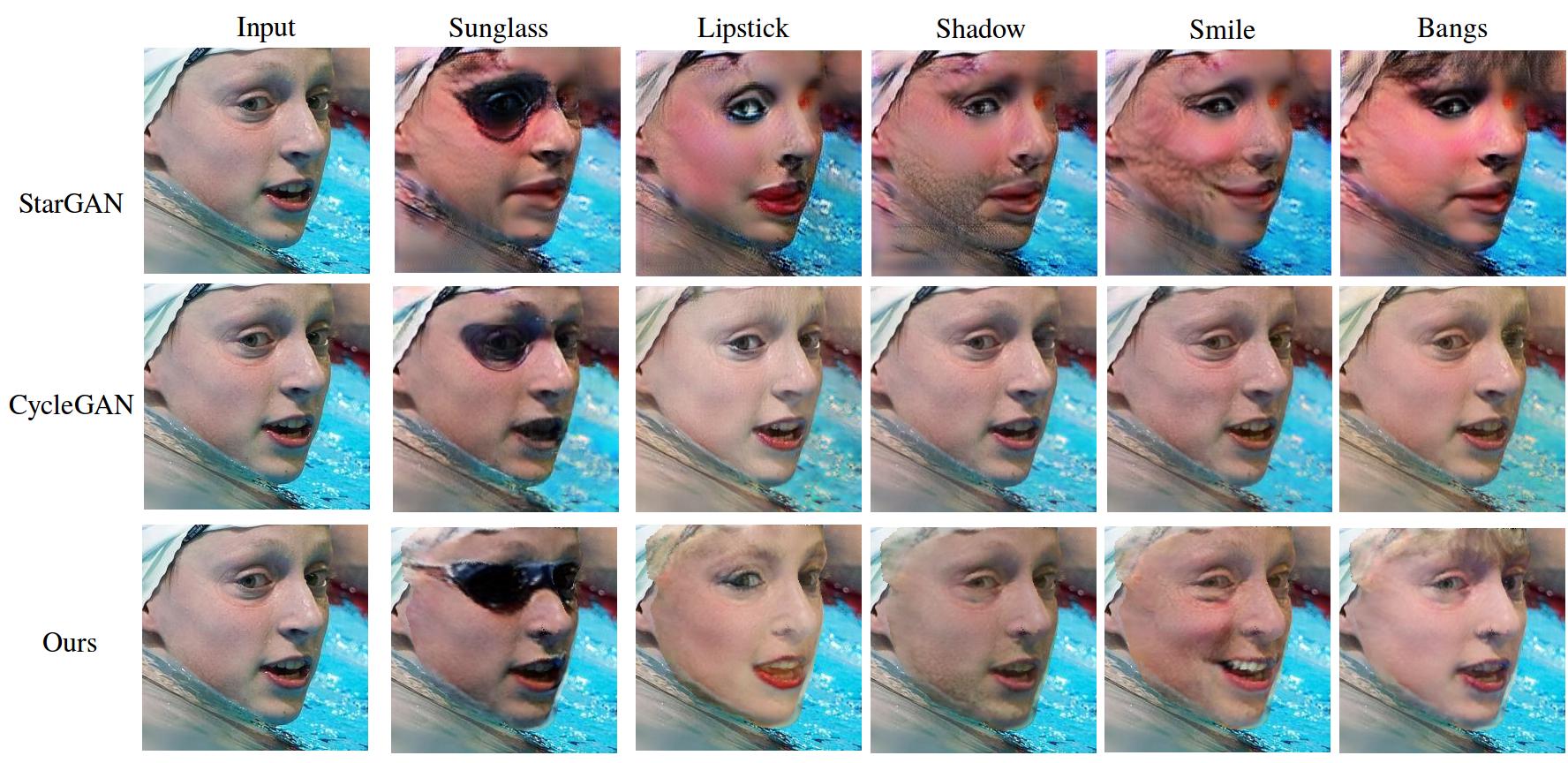}
\caption{Pose-variant qualitative results of our 3DA-GAN compared to StarGAN~\cite{stargan} and CycleGAN~\cite{CycleGAN2017} trained on our prepared data. }
\vspace{-2mm}
\label{fig:attr_visual_5}
\end{figure*}

\begin{figure*}
\centering
\includegraphics[width=1\textwidth]{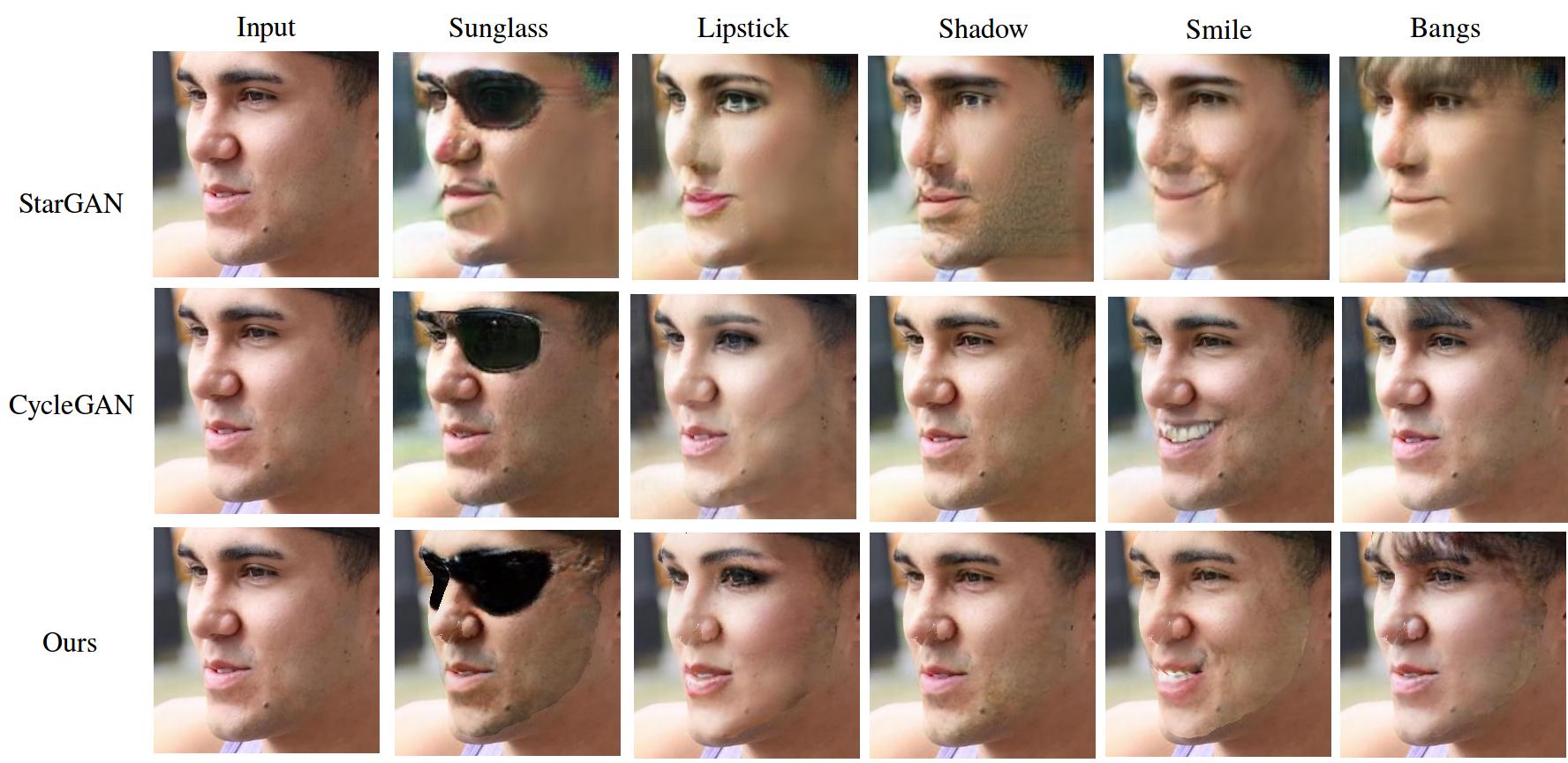}
\caption{Pose-variant qualitative results of our 3DA-GAN compared to StarGAN~\cite{stargan} and CycleGAN~\cite{CycleGAN2017} trained on our prepared data. }
\vspace{-2mm}
\label{fig:attr_visual_6}
\end{figure*}

In addition to Figure 7 and Figure 8 of the main paper, we show more attribute generation results against StarGAN and CycleGAN in Fig.~\ref{fig:attr_visual_2},~\ref{fig:attr_visual_3},~\ref{fig:attr_visual_4}, ~\ref{fig:attr_visual_5}, ~\ref{fig:attr_visual_6}. As can be seen, our method can generate higher quality, more geometrically consistent attributes under large head pose variations.

\begin{figure*}
\centering
\includegraphics[width=1\textwidth]{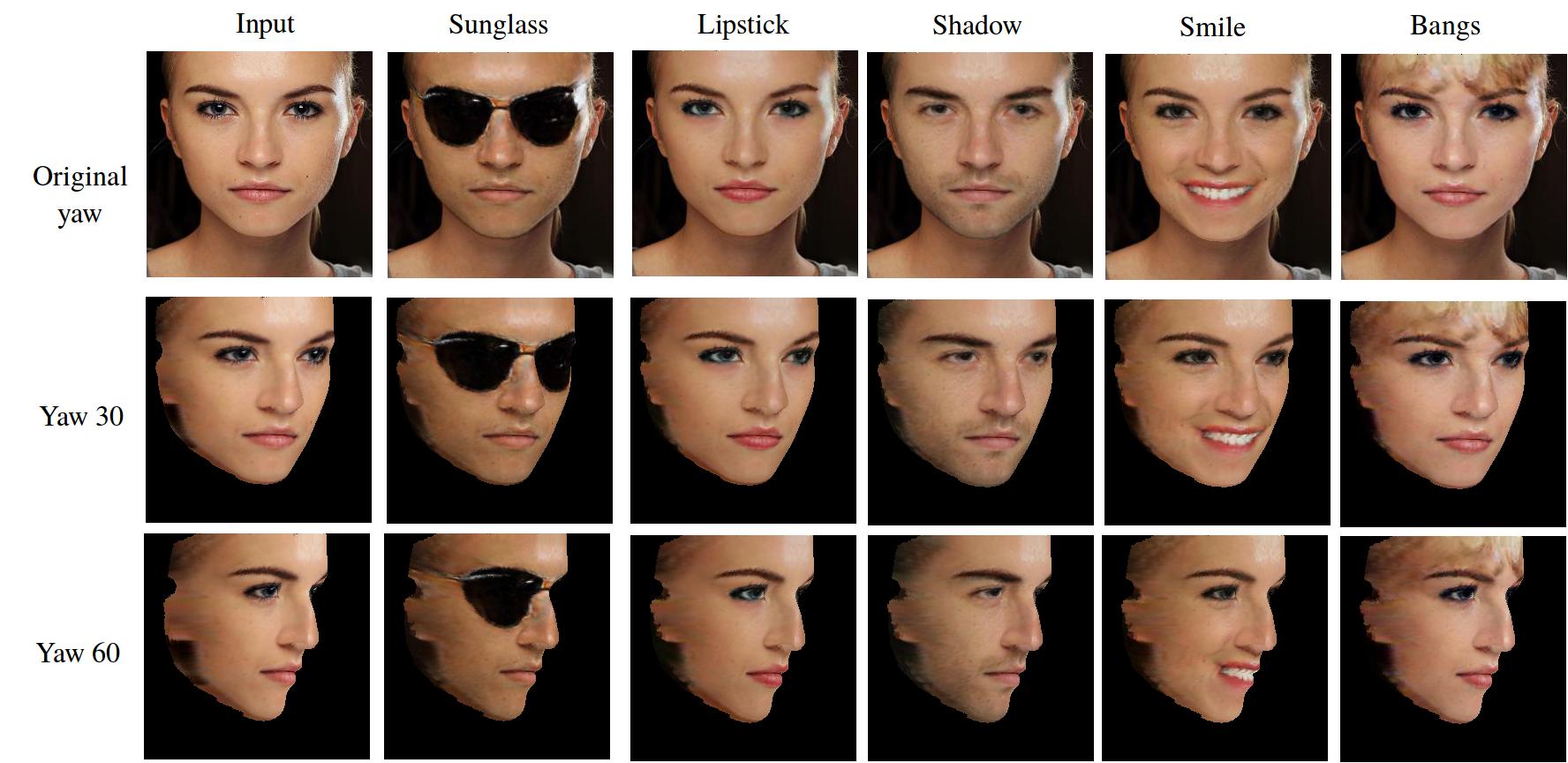}
\caption{Visual results of applying our method to augment face images from CelebA~\cite{celeba} dataset, in attributes and yaw angles.}
\vspace{-2mm}
\label{fig:attr_pose_2}
\end{figure*}

\begin{figure*}
\centering
\includegraphics[width=1\textwidth]{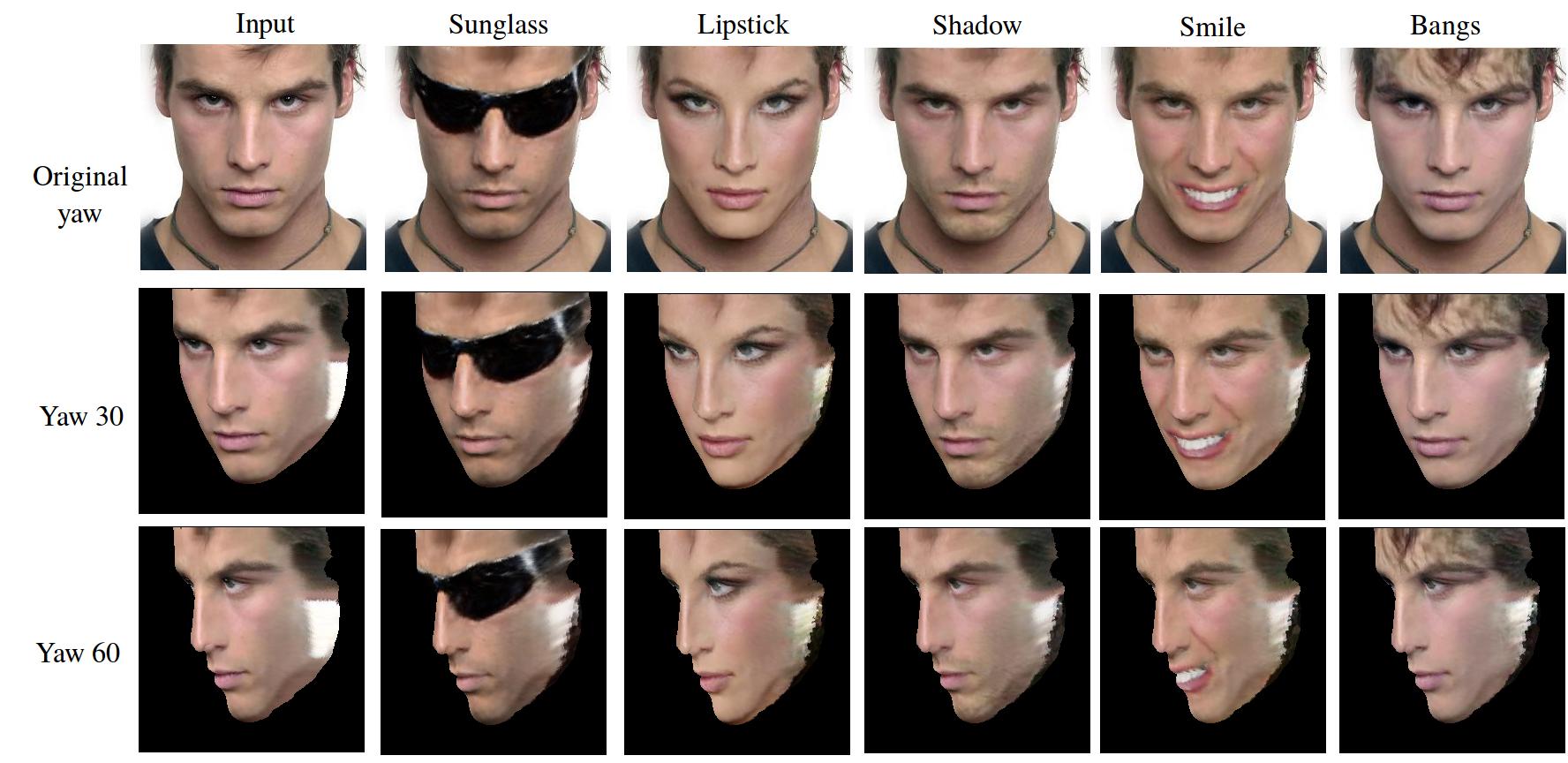}
\caption{Visual results of applying our method to augment face images from CelebA~\cite{celeba} dataset, in attributes and yaw angles.}
\vspace{-2mm}
\label{fig:attr_pose_3}
\end{figure*}

\begin{figure*}
\centering
\includegraphics[width=1\textwidth]{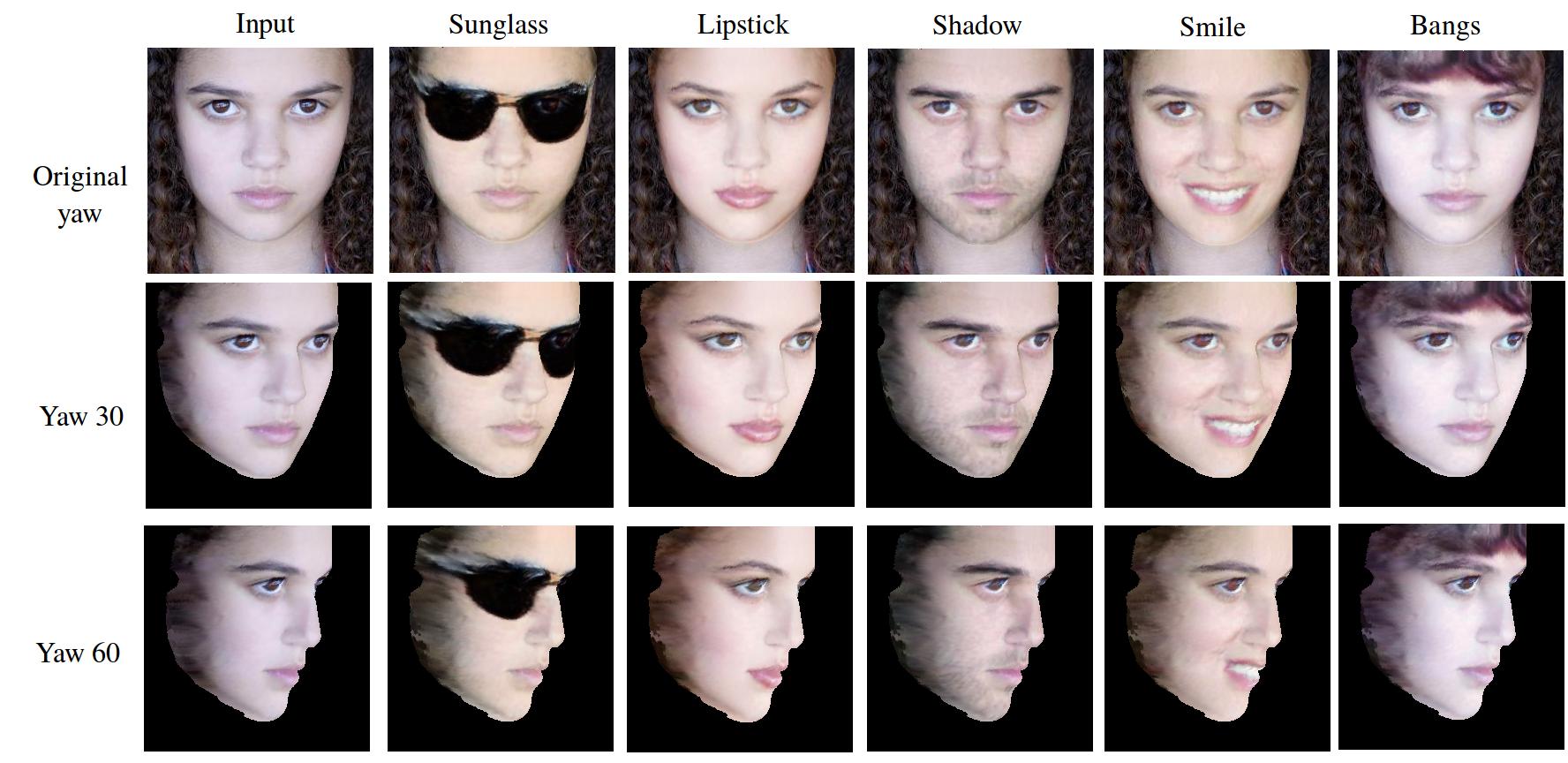}
\caption{Visual results of applying our method to augment face images from CelebA~\cite{celeba} dataset, in attributes and yaw angles.}
\vspace{-2mm}
\label{fig:attr_pose_4}
\end{figure*}

\begin{figure*}
\centering
\includegraphics[width=1\textwidth]{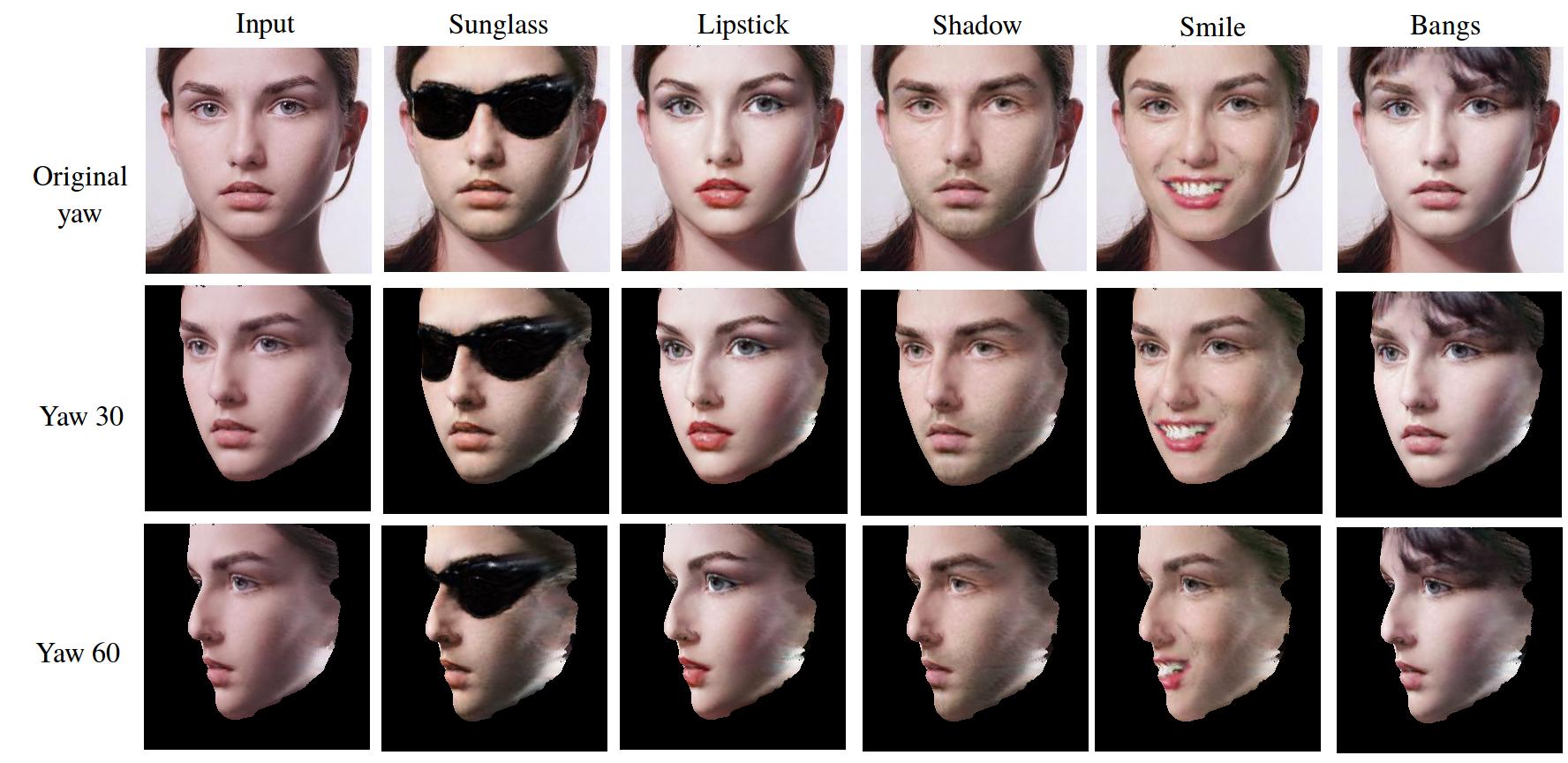}
\caption{Visual results of applying our method to augment face images from CelebA~\cite{celeba} dataset, in attributes and yaw angles.}
\vspace{-2mm}
\label{fig:attr_pose_5}
\end{figure*}

\begin{figure*}
\centering
\includegraphics[width=1\textwidth]{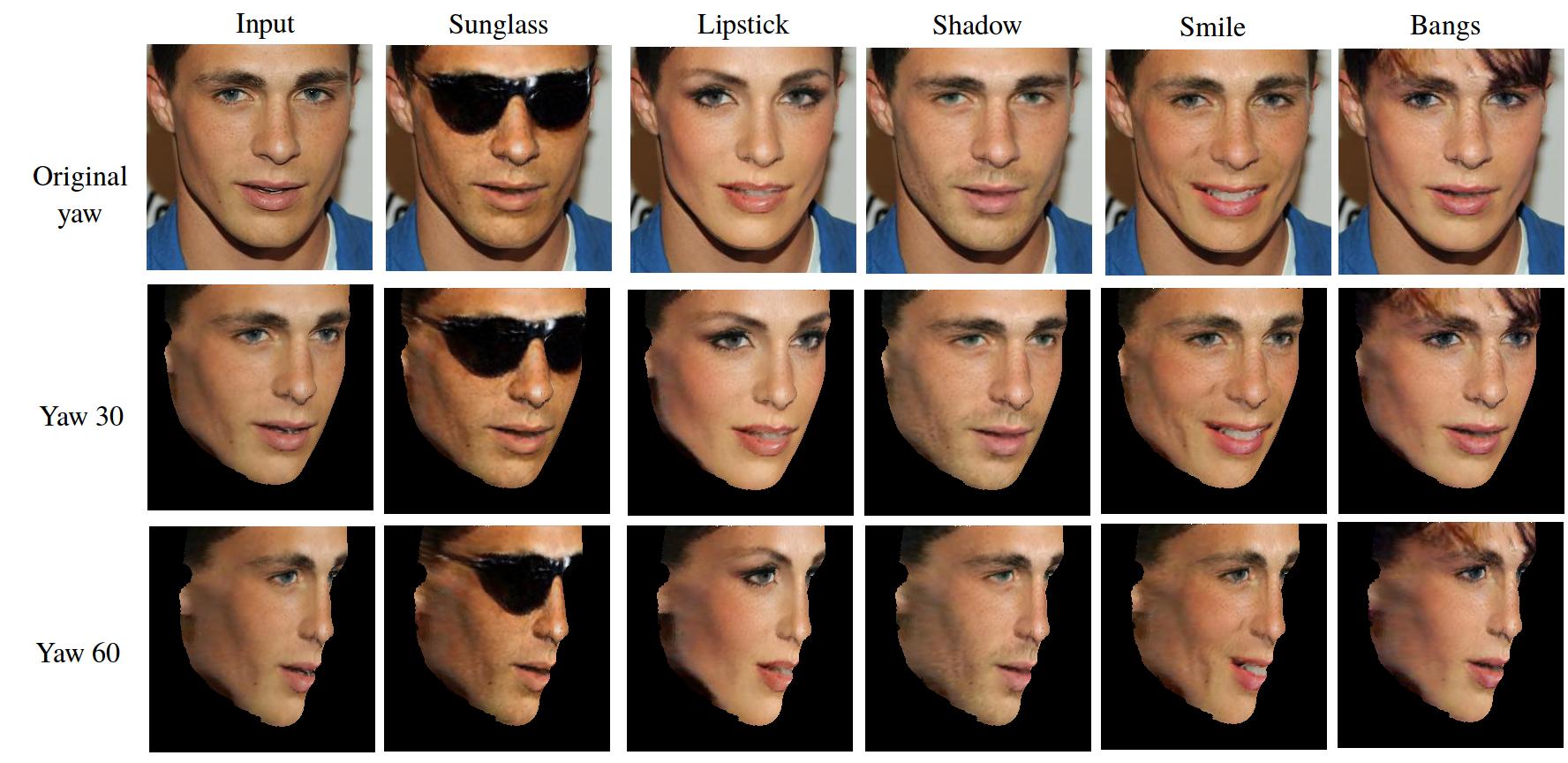}
\caption{Visual results of applying our method to augment face images from CelebA~\cite{celeba} dataset, in attributes and yaw angles.}
\vspace{-2mm}
\label{fig:attr_pose_6}
\end{figure*}

Besides, more attribute and pose augmentation results are shown in Fig.~\ref{fig:attr_pose_2},~\ref{fig:attr_pose_3},~\ref{fig:attr_pose_4},~\ref{fig:attr_pose_5},~\ref{fig:attr_pose_6}. Our method has good potential to benefit the face recognition system by enriching training data diversity while maximumly preserving the original identity information.

\section{Including 5'o Clock Shadow (SH) Results}

In Table 2 of the main paper, we have already shown the attribute classification accuracy and visual quality for the 4 attributes: Sunglasses, Lipstick, Smiling, and Bangs, while omitting the 5'o clock shadow due to the space limit. Therefore, we include the results here for 5'o clock shadow and split the original table into two, one for F1 score in Table~\ref{tab:attr_F1}, and the other for FID score in Table~\ref{tab:attr_FID}. The same trend for SH has shown in both Table~\ref{tab:attr_F1} and Table~\ref{tab:attr_FID}. Our method is consistently better than StarGAN and CycleGAN in the attribute generation accuracy, and achieves consistently lower FID score in the image quality, which indicates more similar visual effects to the original input.

\section{Ablation Study}
\begin{figure*}
\centering
\includegraphics[width=1\textwidth]{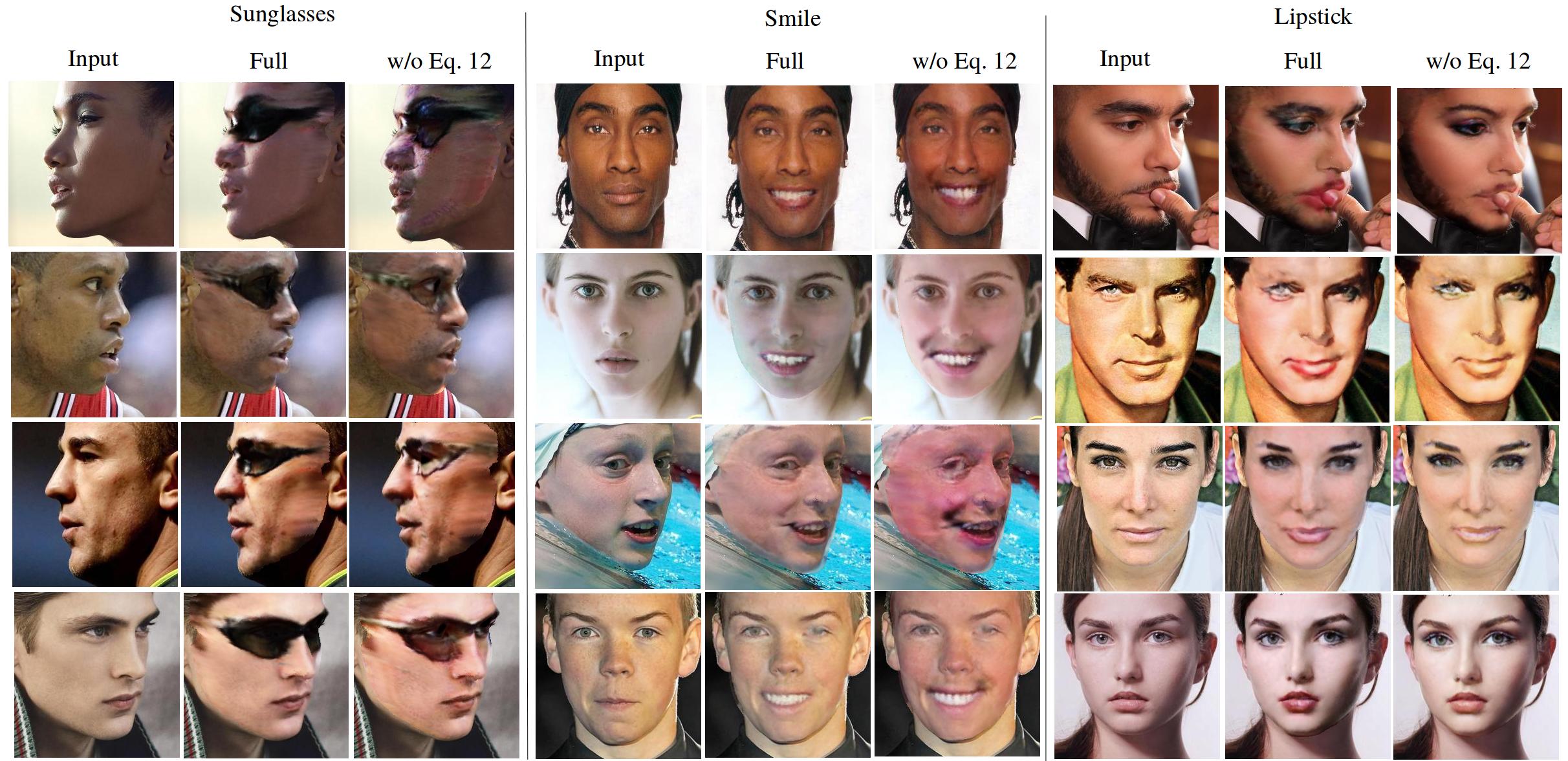}
\caption{The effect of masked reconstruction loss on sunglasses, smile, and lipstick generation.  From left to right:  input images from CelebA dataset, using full losses, without masked reconstruction loss (Eq. 12). The masked reconstruction loss helps generating attributes in a specific region while preserve the non-attribute parts.}
\vspace{-2mm}
\label{fig:maskL}
\end{figure*}

\begin{figure*}
\centering
\includegraphics[width=1\textwidth]{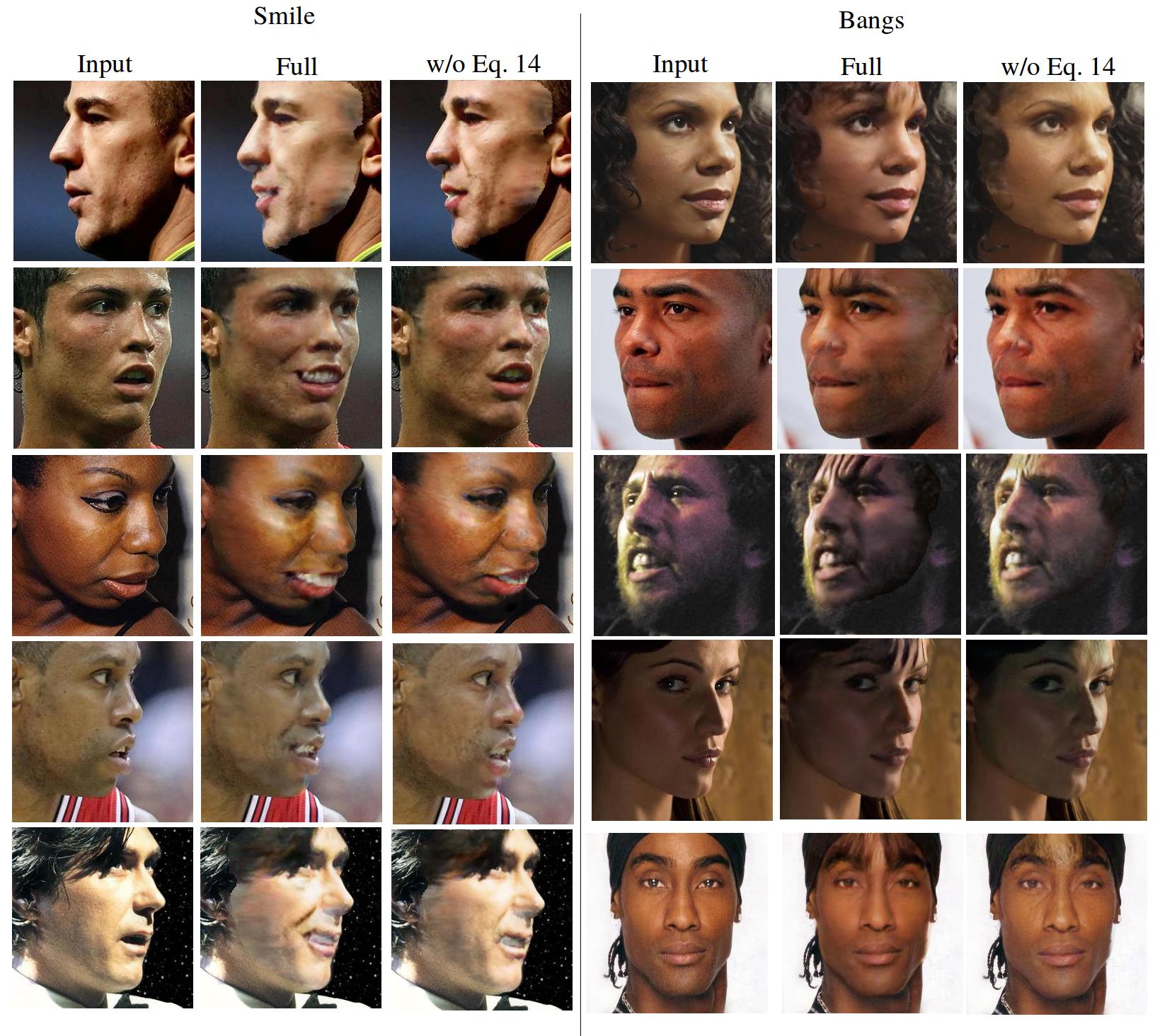}
\caption{The effect of adversarial attribute loss on smile and bangs generation.  From left to right:  input images from CelebA dataset, using full losses, without adversarial attribute loss (Eq. 14). The adversarial attribute loss helps enhancing the intensity of generated attributes.}
\vspace{-2mm}
\label{fig:attrL}
\end{figure*}

\begin{figure}[t]
\centering
\includegraphics[width=0.47\textwidth]{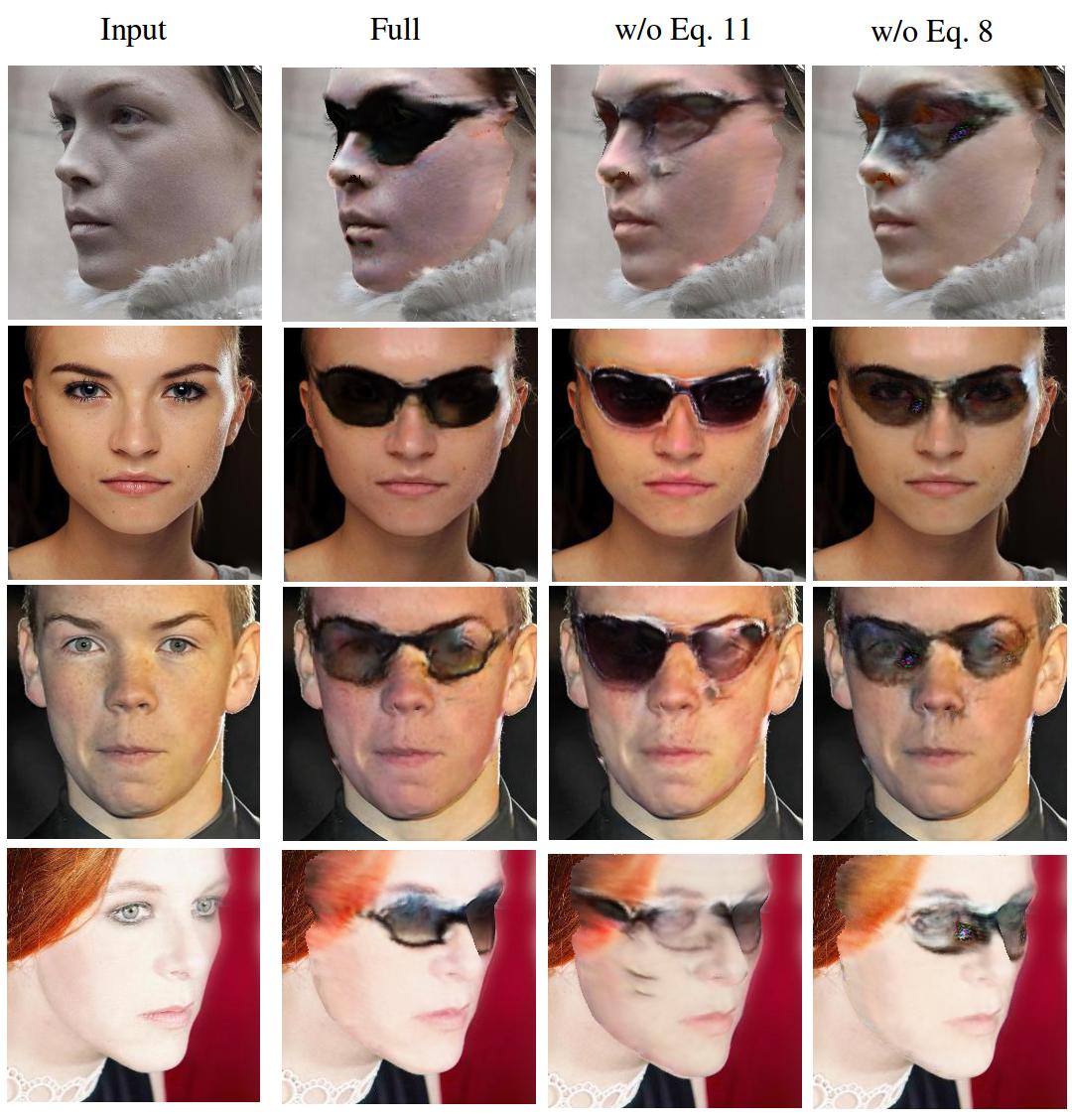}
\caption{The effect of cycle consistent loss and identity loss on sunglasses generation.  From left to right:  input images from CelebA dataset, using full losses, without cycle consistent loss (Eq. 11), and without identity loss (Eq. 8). The cycle consistent loss and identity loss help preserving the non-attribute regions. The identity loss also makes the generated attribute regions more natural.}
\vspace{-2mm}
\label{fig:cycleL}
\end{figure}

Similarly, we include the quantitative ablation study for SH shown in Table~\ref{tab:ablation_F1} and Table~\ref{tab:ablation_FID}. They also show the same trend as other attributes for both F1 score and FID score. More interestingly, we visualize the qualitative generation images by running the ablative models to further indicate the effect of the proposed losses. Figure~\ref{fig:maskL},~\ref{fig:attrL},~\ref{fig:cycleL} show that for ``w/o Eq. 12'', which is masked reconstruction loss, some of the generation fails and some of the generation introduces artifacts. For ``w/o Eq. 14'', which is attribute adversarial loss, the generation mostly fails. For ``w/o Eq. 11'', which is cycle consistency loss, it shows more artifacts than the full results. For ``w/o Eq. 8'', which is identity loss, it also shows certain level of artifact compared to the full result.

\begin{table*}[t]
\begin{center}
\small
\begin{tabular}{@{}l |@{\hskip 1mm} l | c@{\hskip 1mm}c@{\hskip 1mm}c@{\hskip 1mm}c@{\hskip 1mm}c |c@{\hskip 1mm}c@{\hskip 1mm}c@{\hskip 1mm}c@{\hskip 1mm}c}
\hline
\multicolumn{2}{c|}{Test $\to$} &  \multicolumn{5}{c|}{ real (yaw $<$ 45) }  & \multicolumn{5}{c}{ real-a (yaw $\geq$ 45)} \\ \hline %& 
{Train$\downarrow$} & {method$\downarrow$} & SG & LS & SH & SM & BA & SG & LS & SH & SM & BA \\ \hline  
\multirow{3}{*}{StarGAN~\cite{stargan}*} & real & 97.15 & 84.26 & 88.75 & 87.40 & 89.56 & 96.38 & 77.54 & 82.07 & 77.11 & 86.33 \\
& real-a & 97.35 & 78.87 & 89.63 & 83.40  & 89.33 & 98.07 & 75.43 & 88.64 & 79.01 & 86.77 \\
& Ours & \textbf{98.88} & \textbf{84.70} & \textbf{91.12} & \textbf{87.87} & \textbf{94.86} & \textbf{98.23} & \textbf{82.04} & \textbf{90.06} & \textbf{83.32} & \textbf{93.67} \\ \hline 
\multirow{3}{*}{CycleGAN~\cite{CycleGAN2017}*} & real & 97.66 & 84.41 & 84.49 & 86.33  & 70.96  & 90.49 &  74.45 & 79.21 & 76.48  & 69.01 \\ 
& real-a & 98.93 & 91.34 & 85.17 & 84.25  & 82.43  & 97.31 & 69.27 & 84.98 & 75.51 &  80.70 \\
& Ours & \textbf{99.37} & \textbf{94.69} & \textbf{91.80} & \textbf{94.56}  & \textbf{93.35}  & \textbf{99.10} & \textbf{93.04} & \textbf{90.90} & \textbf{91.49} &  \textbf{91.64} \\  \hline 
\end{tabular}
\end{center}
\caption{Quantitative comparison on attribute generation by F1 score on CelebA testing set. The target generated attribute is evaluated by an off-line attribute classifier for F1 score (precision and recall). The higher the better. ``real'' means original CelebA training set.  ``real-a'' means original plus pose augmented images. ``Ours'' means training with our proposed loss and UV texture data. *: we apply the network structure and re-train models. SG: Sunglass, LS: Wearing Lipstick, SH: 5'o clock shadow, SM: Smiling, BA: Bangs.}
\label{tab:attr_F1}
\end{table*}

\begin{table*}[t]
\begin{center}
\small
\begin{tabular}{@{}l |@{\hskip 1mm} l | c@{\hskip 1mm}c@{\hskip 1mm}c@{\hskip 1mm}c@{\hskip 1mm}c |c@{\hskip 1mm}c@{\hskip 1mm}c@{\hskip 1mm}c@{\hskip 1mm}c}
\hline
\multicolumn{2}{c|}{Test $\to$} &  \multicolumn{5}{c|}{ real (yaw $<$ 45) }  & \multicolumn{5}{c}{ real-a (yaw $\geq$ 45)} \\ \hline %& 
{Train$\downarrow$} & {method$\downarrow$} & SG & LS & SH & SM & BA & SG & LS & SH & SM & BA \\ \hline  
\multirow{3}{*}{StarGAN~\cite{stargan}*} & real & 85.68  & 78.86 & 96.97 & 92.28 & 82.28 & 139.77 & 135.93 & 172.84 & 150.58 &  144.02 \\ 
& real-a & 72.73  & 68.91 & 42.36 & 58.92 & 59.53  &  114.02 & 85.14 & 89.02 & 82.82 & 105.34 \\ 
& Ours & \textbf{38.22}  & \textbf{34.05}  & \textbf{26.19} & \textbf{33.02} & \textbf{21.79}  & \textbf{36.31}  & \textbf{35.43} & \textbf{30.05}  & \textbf{30.58} & \textbf{19.39} \\  \hline 
\multirow{3}{*}{CycleGAN~\cite{CycleGAN2017}*} & real & 30.10 & 25.06 & 28.73 & 32.32 & 28.69  & 40.88 & 49.21 & 42.56 & 43.31 & 36.78 \\  
& real-a & 33.89 & \textbf{12.46} & \textbf{6.57} & \textbf{12.74} & \textbf{9.05}  & \textbf{19.83} & 31.04 & 8.81 & 17.06 & 11.54 \\  
& Ours & \textbf{18.54} & \textbf{12.56} & \textbf{7.47} & \textbf{13.03} & \textbf{10.28} & \textbf{29.65} & \textbf{10.92} & \textbf{6.81} & \textbf{10.97}  & \textbf{8.94} \\ \hline 
\end{tabular}
\end{center}
\caption{Quantitative comparison on attribute generation by FID score~\cite{fid2017} on CelebA testing set. Visual quality is indicated by FID score between the target attribute generated images and the ground truth with same attribute images. The lower the better. ``real'' means original CelebA training set.  ``real-a'' means original plus pose augmented images. ``Ours'' means training with our proposed loss and UV texture data. *: we apply the network structure and re-train models. SG: Sunglass, LS: Wearing Lipstick, SH: 5'o clock shadow, SM: Smiling, BA: Bangs.}
\label{tab:attr_FID}
\end{table*}

\begin{table*}[t]
\begin{center}
\small
\begin{tabular}{@{}c |@{\hskip 1.5mm} c | c@{\hskip 1.5mm}c@{\hskip 1.5mm}c@{\hskip 1.5mm}c@{\hskip 1.5mm}c |c@{\hskip 1.5mm}c@{\hskip 1.5mm}c@{\hskip 1.5mm}c@{\hskip 1.5mm}c}
\hline
\multicolumn{2}{c|}{\multirow{2}{*}{Test$\to$}} & \multicolumn{8}{c}{F1-score (higher better)} \\ \cline{3-12}
\multicolumn{2}{c|}{} &  \multicolumn{5}{c|}{ real (yaw $<$ 45) }  & \multicolumn{5}{c}{ real-a (yaw $\geq$ 45)} \\ \hline 
Model & Loss & SG & LS & SH & SM & BA & SG & LS & SH & SM & BA \\ \hline
\multirow{3}{*}{CycleGAN} & w/o 
Eq. 12,14 & 97.97 & 87.92 & 85.05 & 84.62 & 83.65 & 97.93 & 86.21 & 84.40 & 81.11 & 82.21 \\
 & w/o Eq. 12 & 99.28 & 92.95 & 90.10 & 93.17 & \textbf{94.86} & 98.87 & \textbf{90.79} & 89.15 & 89.50 & \textbf{93.82} \\
(ResNet) & w/o Eq. 14 & 97.82 & 83.28 & 82.25 & 81.81 & 86.56 & 97.54 & 82.35 & 82.58 & 78.43 & 85.86 \\
 & Full & \textbf{99.37} & \textbf{94.69} & \textbf{91.80} & \textbf{94.56}  & \textbf{93.35}  & \textbf{99.10} & \textbf{93.04} & \textbf{90.90} & \textbf{91.49} & \textbf{91.64} \\ \hline 
\end{tabular}
\end{center}
\caption{Ablation study for w/o masked reconstruction loss (Eq. 12)), and/or w/o attribute loss (Eq. 14). F1 scores are reported. We use CycleGAN ResNet structure as it achieves the best result across the experiments. SG: Sunglass, LS: Wearing Lipstick, SH: 5'o clock shadow, SM: Smiling, BA: Bangs.} 
\label{tab:ablation_F1}
\vspace{-3mm}
\end{table*}

\begin{table*}[t]
\begin{center}
\small
\begin{tabular}{@{}c |@{\hskip 1.5mm} c | c@{\hskip 1.5mm}c@{\hskip 1.5mm}c@{\hskip 1.5mm}c@{\hskip 1.5mm}c |c@{\hskip 1.5mm}c@{\hskip 1.5mm}c@{\hskip 1.5mm}c@{\hskip 1.5mm}c}
\hline
\multicolumn{2}{c|}{\multirow{2}{*}{Test$\to$}} & \multicolumn{8}{c}{FID-score (lower better)} \\ \cline{3-12}
\multicolumn{2}{c|}{} &  \multicolumn{5}{c|}{ real (yaw $<$ 45) }  & \multicolumn{5}{c}{ real-a (yaw $\geq$ 45)} \\ \hline 
Model & Loss & SG & LS & SH & SM & BA & SG & LS & SH & SM & BA \\ \hline
\multirow{3}{*}{CycleGAN} & w/o 
Eq. 12,14 & 20.2 & \textbf{10.6} & 13.9 & \textbf{7.8} & 14.1  & 43.8 & 20.4 & 20.4 & 27.2 & 18.3 \\
 & w/o Eq. 12 & \textbf{17.6} & 17.5 & \textbf{7.0} & 13.8  & 11.9  & \textbf{26.6} & 18.1 & 11.4 & 15.0 & 11.4 \\
(ResNet) & w/o Eq. 14 & 29.1 & 19.0 & 7.6 & 18.1 & \textbf{10.5} & 39.3 & 18.4 & 11.3 & 17.7 & 10.4 \\
 & Full & \textbf{18.5} & \textbf{12.6} & \textbf{7.5} & \textbf{13.0} & \textbf{10.3} & \textbf{29.7} & \textbf{10.9} & \textbf{6.8} & \textbf{11.0}  & \textbf{8.9} \\ \hline
\end{tabular}
\end{center}
\caption{Ablation study for w/o masked reconstruction loss (Eq. 12)), and/or w/o attribute loss (Eq. 14). FID scores are reported. We use CycleGAN ResNet structure as it achieves the best result across the experiments. SG: Sunglass, LS: Wearing Lipstick, SH: 5'o clock shadow, SM: Smiling, BA: Bangs.} 
\label{tab:ablation_FID}
\vspace{-3mm}
\end{table*}

\section{Network Architectures}

The network architectures of StarGAN and CycleGAN used in our experiments are shown in Table~\ref{tab:G_stargan},~\ref{tab:DQ_DA_stargan},~\ref{tab:G_cyclegan},~\ref{tab:DQ_cyclegan},~\ref{tab:DA_cyclegan}. We use instance normalization for the generator network in all the layers except the output layer. For the quality and attribute discriminator networks, we use Leaky ReLU with a negative slope of 0.01 in StarGAN, and 0.02 in CycleGAN. The definitions of the annotations in the tables are as follows: C: the number of output channels, K: kernel size, S: stride size, P: padding size, IN: instance normalization, $A_d$: the number of attributes to be generated, $h$ and $w$ are the height and width of the input image.

\begin{table}[t]
\begin{center}
\small
\begin{tabular}{l|c} \hline
Type & Layer \\ \hline 
Downsampling & Conv-(C64, K7x7, S1, P3), IN, ReLU \\
Downsampling & Conv-(C128, K4x4, S2, P1), IN, ReLU \\
Downsampling & Conv-(C256, K4x4, S2, P1), IN, ReLU \\ \hline
Residual Block & Conv-(C256, K3x3, S1, P1), IN, ReLU \\ 
Residual Block & Conv-(C256, K3x3, S1, P1), IN, ReLU \\
Residual Block & Conv-(C256, K3x3, S1, P1), IN, ReLU \\ 
Residual Block & Conv-(C256, K3x3, S1, P1), IN, ReLU \\
Residual Block & Conv-(C256, K3x3, S1, P1), IN, ReLU \\ 
Residual Block & Conv-(C256, K3x3, S1, P1), IN, ReLU \\ \hline
Upsampling & Deconv-(C128, K4x4, S2, P1), IN, ReLU \\ 
Upsampling & Deconv-(C64, K4x4, S2, P1), IN, ReLU \\ 
Upsampling & Deconv-(C3, K7x7, S1, P3), Tanh \\ \hline
\end{tabular}
\end{center}
\caption{StarGAN Generator network architecture}
\vspace{-2mm}
\label{tab:G_stargan}
\vspace{-2mm}
\end{table}

\begin{table}[t]
\begin{center}
\small
\begin{tabular}{l|c} \hline
Type & Layer \\ \hline 
Input & Conv-(C64, K4x4, S2, P1), Leaky ReLU \\ \hline
Hidden & Conv-(C128, K4x4, S2, P1), Leaky ReLU \\
Hidden & Conv-(C256, K4x4, S2, P1), Leaky ReLU \\
Hidden & Conv-(C512, K4x4, S2, P1), Leaky ReLU \\
Hidden & Conv-(C1024, K4x4, S2, P1), Leaky ReLU \\
Hidden & Conv-(C2048, K4x4, S2, P1), Leaky ReLU \\ \hline
Output & Conv-(C1, K3x3, S1, P1) \\
Output & Conv-(C$A_d$, $K\frac{h}{64}\times\frac{w}{64}$, S1, P0) \\ \hline
\end{tabular}
\end{center}
\caption{StarGAN Quality and Attribute discriminator network architecture}
\vspace{-2mm}
\label{tab:DQ_DA_stargan}
\vspace{-2mm}
\end{table}

\begin{table}[t]
\begin{center}
\small
\begin{tabular}{l|c} \hline
Type & Layer \\ \hline 
Input & ReflectionPad2d(3) \\
Input & Conv-(C64, K7x7, S1, P0), IN, ReLU \\ \hline
Downsampling & Conv-(C128, K3x3, S2, P1), IN, ReLU \\
Downsampling & Conv-(C256, K3x3, S2, P1), IN, ReLU \\ \hline
Residual Block & Conv-(C256, K3x3, S1, P0), IN, ReLU \\
Residual Block & Conv-(C256, K3x3, S1, P0), IN, ReLU \\
Residual Block & Conv-(C256, K3x3, S1, P0), IN, ReLU \\
Residual Block & Conv-(C256, K3x3, S1, P0), IN, ReLU \\
Residual Block & Conv-(C256, K3x3, S1, P0), IN, ReLU \\
Residual Block & Conv-(C256, K3x3, S1, P0), IN, ReLU \\
Residual Block & Conv-(C256, K3x3, S1, P0), IN, ReLU \\
Residual Block & Conv-(C256, K3x3, S1, P0), IN, ReLU \\
Residual Block & Conv-(C256, K3x3, S1, P0), IN, ReLU \\ \hline
Upsampling & Deconv-(C128, K3x3, S2, P1), IN, ReLU \\
Upsampling & Deconv-(C64, K3x3, S2, P1), IN, ReLU \\
Upsampling & ReflectionPad2d(3) \\
Upsampling & Deconv-(C3, K7x7, S1, P0), Tanh \\ \hline
\end{tabular}
\end{center}
\caption{CycleGAN Generator network architecture}
\vspace{-2mm}
\label{tab:G_cyclegan}
\vspace{-2mm}
\end{table}

\begin{table}[t]
\begin{center}
\small
\begin{tabular}{l|c} \hline
Type & Layer \\ \hline 
Input & Conv-(C64, K4x4, S2, P1), Leaky ReLU \\ \hline
Hidden & Conv-(C128, K4x4, S2, P1), Leaky ReLU \\
Hidden & Conv-(C256, K4x4, S2, P1), Leaky ReLU \\
Hidden & Conv-(C512, K4x4, S1, P1), Leaky ReLU \\ \hline
Output & Conv-(C1, K4x4, S1, P1) \\ \hline
\end{tabular}
\end{center}
\caption{CycleGAN quality discriminator network architecture}
\vspace{-2mm}
\label{tab:DQ_cyclegan}
\vspace{-2mm}
\end{table}

\begin{table}[t]
\begin{center}
\small
\begin{tabular}{l|c} \hline
Type & Layer \\ \hline 
Input & Conv-(C64, K4x4, S2, P1), Leaky ReLU \\ \hline
Hidden & Conv-(C128, K4x4, S2, P1), Leaky ReLU \\
Hidden & Conv-(C256, K4x4, S2, P1), Leaky ReLU \\
Hidden & Conv-(C512, K4x4, S2, P1), Leaky ReLU \\
Hidden & Conv-(C1024, K4x4, S2, P1), Leaky ReLU \\
Hidden & Conv-(C2048, K4x4, S2, P1), Leaky ReLU \\ \hline
Output & Conv-(C$A_d$, $K\frac{h}{64}\times\frac{w}{64}$, S1, P0) \\ \hline
\end{tabular}
\end{center}
\caption{The Attribute discriminator network architecture we used with CycleGAN}
\vspace{-2mm}
\label{tab:DA_cyclegan}
\vspace{-2mm}
\end{table}

\end{document}